\documentclass{article}

\usepackage{arxiv}

\usepackage{newtxtext}
\usepackage{newtxmath}

\usepackage[utf8]{inputenc} % allow utf-8 input
\usepackage[T2A,T1]{fontenc}    % support for Cyrillic (T2A) and Latin (T1)
\usepackage[russian,english]{babel} % hyphenation rules for Russian and English
\usepackage{hyperref}       % hyperlinks
\usepackage{url}            % simple URL typesetting
\usepackage{booktabs}       % professional-quality tables
\usepackage{amsfonts}       % blackboard math symbols
\usepackage{nicefrac}       % compact symbols for 1/2, etc.
\usepackage{microtype}      % microtypography
\usepackage{cleveref}       % smart cross-referencing
\usepackage{lipsum}         % Can be removed after putting your text content
\usepackage{graphicx}
\usepackage{natbib}
\usepackage{doi}

\newif\ifuniqueAffiliation
\uniqueAffiliationfalse   % set to true for single affiliation

\title{A Comparative Analysis of Machine Learning Algorithms for Multi-Task Prediction of the Parameters of the Pectin Hydrolysis–Extraction Process}

\ifuniqueAffiliation
    \author{%
        \href{https://orcid.org/0000-0003-2525-1183}{\includegraphics[scale=0.06]{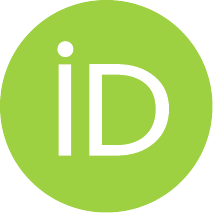}\hspace{1mm}M. K. Arabov}\thanks{Email: \texttt{MKArabov@kpfu.ru}} \\
        Institute of Computational Mathematics and Information Technologies\\
        Kazan Federal University\\
        Kazan, Russia \\
        \texttt{MKArabov@kpfu.ru}
    }
\else
    \usepackage{authblk}
    
    \newbox{\orcid}\sbox{\orcid}{\includegraphics[scale=0.06]{orcid.pdf}}
    \author[1]{%
        \href{https://orcid.org/0000-0003-2525-1183}{\usebox{\orcid}\hspace{1mm}M. K. Arabov}\thanks{Email: \texttt{MKArabov@kpfu.ru}}%
    }
    \author[2]{%
        Sh. Yo. Kholov\thanks{Email: \texttt{shavkat.kholov@yandex.ru}}%
    }
    \author[3]{%
        Z. K. Muhiddin\thanks{Email: \texttt{zainy@mail.ru}}%
    }
    \affil[1]{Institute of Computational Mathematics and Information Technologies, Kazan Federal University, Kazan, Russia}
    \affil[2]{Tajik Technical University named after Academician M.S. Osimi, Dushanbe, Tajikistan}
    \affil[3]{V.I. Nikitin Institute of Chemistry, National Academy of Sciences of Tajikistan, Dushanbe, Tajikistan}
\fi

\renewcommand{\shorttitle}{ML for Pectin Process Prediction}

\hypersetup{
    pdftitle={A Comparative Analysis of Machine Learning Algorithms for Multi-Task Prediction of Pectin Hydrolysis–Extraction Process Parameters},
    pdfsubject={cs.LG},
    pdfauthor={M. K. Arabov, Sh. Yo. Kholov, Z. K. Muhiddin},
    pdfkeywords={machine learning, multi-task regression, gradient boosting, pectin, flash hydrolysis, SHAP, interpretable AI, process optimisation, statistical analysis, feature mapping},
}

\begin{document}
\maketitle

\begin{abstract}
This study addresses the challenge of controlling a complex, multi-parameter technological process---pectin hydrolysis--extraction---using machine learning methods. The experimental foundation of the work is a unique database comprising 1,000 laboratory experiments conducted under controlled conditions on real process equipment. The trials encompass seven types of plant raw material (apple pomace of two varieties, quince, apricot, sunflower head, rhubarb, and pumpkin) and four variable process factors (temperature $85$--$130\,^{\circ}\mathrm{C}$, pressure $0.9$--$2.2$~atm, holding time $3$--$10$~min, pH $1.5$--$2.0$). Four output characteristics were recorded for each experiment: pectin yield, galacturonic acid content, molecular weight, and degree of esterification.

During the data preprocessing stage, a logarithmic transformation of time was performed (to account for distribution asymmetry), categorical features were encoded (raw material type, extraction method---flash/traditional), missing values were imputed with medians, and the data were standardised. A unified mapping of Russian and English variable names was established to ensure analytical reproducibility. An extensive statistical analysis was conducted, including Analysis of Variance (ANOVA) by raw material type for each target variable (all differences are statistically significant, $p < 0.05$), a pairwise comparison of extraction methods with a preliminary normality check, and a correlation matrix with an assessment of coefficient significance.

The scientific novelty of the work lies in the development of a methodology for the comparative analysis of machine learning algorithms for multi-task regression under real technological process conditions, utilising a large-scale experimental database. To solve the multi-task regression problem, 11 algorithms representing the main machine learning paradigms were trained and compared: regularised linear models (Ridge, Lasso, ElasticNet), ensemble methods (Random Forest, Gradient Boosting, XGBoost, CatBoost, Extra Trees), the $k$-nearest neighbours method ($K$ Neighbors), the support vector method (SVR), and a multilayer perceptron (MLP). Simple linear regression and LightGBM were excluded from the final comparison due to low efficiency and feature multicollinearity. All models were optimised using randomised hyperparameter search within cross-validation. The interpretability of the solutions was ensured by calculating SHAP values for each target variable and constructing partial dependence plots.

The best results were demonstrated by the CatBoost algorithm (average coefficient of determination $R^{2} \approx 0.946$ after hyperparameter optimisation), confirming the effectiveness of gradient boosting on categorical features. Feature importance analysis revealed the dominant role of the raw material type, as well as a significant influence of hydrolysis duration (on a logarithmic scale) and temperature. A technological schedule for selecting optimal conditions to maximise pectin yield and quality is proposed.

The developed preprocessing and prediction pipeline was exported in a production-ready format, preserving all artefacts (models, scalers, encoders, metadata, feature correspondence map). The capability for batch prediction on new experimental data with automatic alignment to a unified feature structure was demonstrated.

The practical significance of the work lies in the creation of a methodology for the objective selection of machine learning algorithms for multi-task prediction under the conditions of real technological processes for processing plant raw materials. It is shown that the application of ensemble methods, combined with rigorous statistical analysis and interpretable artificial intelligence, not only allows for a significant reduction in the volume of necessary physical experiments but also forms the basis for creating an intelligent pectin production control system.
\end{abstract}


\section{Introduction}
\label{sec:introduction}

Multi-parameter technological processes remain difficult to control when numerous inputs non-linearly affect several key product properties, and traditional optimisation demands expensive, large-scale experimentation~\citep{friedman2001greedy,breiman2001random}.  
The present study focuses on flash hydrolysis--extraction of pectin, a process governed by temperature, pressure, treatment time, medium acidity, and raw material type, and yielding four interdependent output characteristics~\citep{fishman2003flash,kholov2017modeling}.  
Pectin is a high-value polysaccharide widely used as a gelling agent, thickener, and stabiliser in the food, pharmaceutical, and cosmetic industries~\citep{jafari2017pectin,riyamol2023recent}.  
The strong non-linearity and the decisive role of categorical features (especially plant raw material) make conventional modelling insufficient~\citep{santosh2023current,barrera2025comprehensive}.

Machine learning (ML), and in particular multi-task learning, offers a promising route for simultaneous prediction of correlated output parameters~\citep{ruder2017overview}.  
Although numerous algorithms have been applied in various industrial contexts~\citep{shahani2022machine,cha2022hybrid,sudarshan2025advancing}, no comprehensive comparison of methodologically diverse ML families exists for multi-task regression on real industrial pectin data.  
Equally under-explored are the interpretability of the obtained models and the rigorous statistical validation of the chosen algorithm.

The scientific novelty of this work consists in a methodology for systematic comparative analysis of ML algorithms for multi-task prediction of pectin flash-hydrolysis parameters.  
On the basis of a uniquely large experimental database (1,000 controlled laboratory trials), eleven algorithms spanning regularised linear models, ensemble methods, support vector regression, k-nearest neighbours, and a multilayer perceptron were benchmarked.  
For the first time, the comparison is complemented by SHAP-based interpretability analysis and extended statistical testing (ANOVA for raw-material significance, Mann--Whitney test for extraction methods).  

The aim is to develop and experimentally validate a methodology for selecting optimal ML algorithms for multi-task prediction in plant-raw-material processing, and to deliver a practical real-time tool.  
The registered computer programme ``PectinProductionPredicator''~\citep{arabov2026pectinproductionpredicator} provides the trained models through an interactive web interface~\citep{arabovsailab2024pectinmodels}.

\section{Literature Review}
\label{sec:literature-review}

\subsection*{Machine learning in process prediction and the need for systematic comparison}
Multi-task learning has been shown to exploit correlations among output variables, improving accuracy over single-output models~\citep{ruder2017overview}.  
In the domain of pectin processing, several recent studies illustrate this potential but also reveal fragmented coverage of algorithms.  
Siejak et al.~\citep{siejak2024prediction} predicted pectin solution viscosity using a limited set of models; the work demonstrated the feasibility of ML but did not extend to multi-task outputs nor include tree-based ensembles with categorical handling.  
Yapias et al.~\citep{yapias2025optimized} optimised pectin extraction from various raw materials, yet the comparison was restricted to a narrow class of methods, leaving open the question of whether substantially different algorithm families could offer better trade-offs between accuracy and interpretability.  
Fan et al.~\citep{fan2022automated} proposed automated hyperparameter tuning for gradient boosting in geological exploration and suggested its transfer to technological tasks, but did not perform a head-to-head benchmark of diverse algorithms on a unified pectin dataset.  

These works, together with broader industrial ML surveys~\citep{sudarshan2025advancing}, confirm that a systematic multi-algorithm comparison under controlled conditions is still missing, especially for multi-task regression problems where the data include both numerical and categorical predictors.

\subsection*{Ensemble methods and categorical features}
Gradient boosting~\citep{friedman2001greedy} and random forests~\citep{breiman2001random} form the backbone of modern ensemble learning.  
Subsequent implementations---XGBoost, LightGBM, and CatBoost---extended these ideas, each introducing specific optimisations~\citep{pedregosa2011scikit}.  
CatBoost, in particular, incorporates an ordered boosting mechanism and native handling of categorical features via target-based statistics, making it especially attractive for datasets where raw material type is a dominant categorical predictor~\citep{prokhorenkova2018catboost}.  
Hua et al.~\citep{hua2024novel} applied CatBoost to industrial quality estimation and used SHAP values to extract key process variables, demonstrating that high predictive performance can coexist with interpretability in real production settings.  
Zhou et al.~\citep{zhou2024novel} further confirmed the promise of ML for co-extraction of pectin and essential oils, albeit without a broad algorithm comparison.

\subsection*{Interpretability and statistical validation}
The ``black-box'' nature of many ML models has motivated the development of explainable artificial intelligence (XAI).  
SHAP (SHapley Additive exPlanations) decomposes a prediction into additive feature contributions grounded in game theory~\citep{lundberg2017unified}, while LIME builds local surrogate models for individual predictions~\citep{ribeiro2016why}.  
Despite their power, these tools have rarely been applied to multi-task pectin prediction, and no previous study has combined SHAP analysis with formal statistical tests (ANOVA for categorical factors, Mann--Whitney for process variants) to validate both the model behaviour and the significance of technological parameters.

Practical deployment of ML in resource-constrained industrial settings often requires compact yet accurate models; hybrid and Tiny ML approaches have been explored for such engineering problems~\citep{vuppalapati2021crossing}.  
This further motivates the need for a methodology that not only identifies the most accurate algorithm but also considers interpretability and resource efficiency.

\subsection*{Positioning of the present study}
The reviewed literature indicates that:
\begin{itemize}
    \item existing pectin-related ML works are limited in the breadth of algorithms compared, in the number of simultaneously predicted outputs, or in the scale of experimental data;
    \item the unique ability of algorithms such as CatBoost to exploit categorical structure has not been evaluated on a large hydrolysis--extraction dataset;
    \item XAI methods and rigorous statistical tests have not been integrated into a unified methodology for this application.
\end{itemize}
The present study closes these gaps by:
\begin{itemize}
    \item benchmarking 11 algorithm families on 1,000 controlled laboratory trials with four multi-task outputs;
    \item explicitly contrasting algorithms that natively handle categorical features (e.g., CatBoost) with those that require preprocessing;
    \item providing a complete interpretability layer via SHAP and validating the influence of raw material type and extraction method through ANOVA and Mann--Whitney tests.
\end{itemize}

\section{Materials and Methods}
\label{sec:materials-and-methods}
\subsection{Experimental Data}

Experimental investigations were carried out on seven types of plant raw material containing pectin: apple pomace of two varieties (Faizabad variety---AP(F) and Muminabad variety---AP(M)), quince (Qnc.), apricot (Apr.), sunflower heads (SFH), rhubarb (Rhb.), and pumpkin (Pmp.). The raw material was pre-ground to a particle size of $0.5$--$1$~mm and normalised by moisture content to $10$--$15\%$ to ensure process reproducibility. Pectin extraction was performed using flash hydrolysis technology in an autoclave with the following control parameters being varied: temperature in the range $85$--$130\,^{\circ}\mathrm{C}$ (main regimes $110$--$130\,^{\circ}\mathrm{C}$, individual trials at $85\,^{\circ}\mathrm{C}$ for the traditional method), pressure $0.9$--$2.2$~atm, holding time $3$--$10$~min (for the flash method). The acidity of the medium was maintained in the pH interval $1.5$--$2.0$ (working range $1.8$--$2.0$). Upon completion of the holding period, an instantaneous pressure release was performed, creating a steam-explosion effect that promotes the depolymerisation of protopectin and increases the yield of the target product. The research was conducted in the Laboratory of High-Molecular Compounds Chemistry of the V.I. Nikitin Institute of Chemistry of the National Academy of Sciences of Tajikistan during the period 2012--2024. Over the specified period, 1,000 trials were conducted, which formed the experimental database. Table~\ref{tab:experimental-data} presents a fragment of the obtained experimental data (the first five records), illustrating the structure of the initial sample.

\begin{table}[htbp]
\centering
\caption{Experimental data of the pectin hydrolysis--extraction process (fragment)}
\label{tab:experimental-data}
\begin{tabular}{cccccccccc}  % 10 столбцов (было 9)
\toprule
\textbf{No.} & \textbf{Pectin sample} & \textbf{$t$, min} & \textbf{$T$, $^{\circ}$C} & \textbf{$P$, atm} & \textbf{pH} & \textbf{PY, \%} & \textbf{GA, \%} & \textbf{$M_{w}$, Da} & \textbf{DE, \%} \\
\midrule
1 & AP(M) & 7 & 120 & 2.08 & 2.00 & 25.864 & 52.706 & 103773.64 & 71.17 \\
2 & AP(M) & 7 & 120 & 1.74 & 2.08 & 24.830 & 51.645 & 103098.49 & 70.02 \\
3 & Apr. & 5 & 130 & 2.09 & 1.74 & 14.755 & 67.550 & 127235.35 & 82.81 \\
4 & AP(M) & 7 & 120 & 2.05 & 2.00 & 26.353 & 53.804 & 105994.85 & 65.42 \\
5 & SFH & 10 & 110 & 1.03 & 2.00 & 19.505 & 66.606 & 145498.37 & 67.76 \\
\bottomrule
\end{tabular}
\end{table}

The analytical characterisation of the obtained pectin fractions included gravimetric determination of pectin yield (PY, \%), titrimetric determination of the mass fraction of galacturonic acid (GA, \%), determination of the weight-average molecular weight by high-performance liquid chromatography (HPLC, $M_{w}$, Da), and titrimetric determination of the degree of esterification (DE, \%). In total, five input parameters (temperature, pressure, holding time, acidity of the medium, type of raw material) and four output characteristics (PY, GA, $M_{w}$, DE) were recorded for each experiment.

\subsection{Data Preprocessing}

Data preprocessing was performed according to the following scheme. First, auxiliary and completely empty columns were removed, after which the Russian-language column names were renamed to English-language ones with the formation of a correspondence map to ensure analytical reproducibility. Categorical features were transformed: the type of raw material was encoded with integer labels (label encoding), and the binary feature ``extraction method'' was introduced as $\text{method\_encoded} = 1$ for holding time $\leq 15$~min (flash method) and $\text{method\_encoded} = 0$ otherwise (traditional method). Due to the high asymmetry of the holding time distribution (the skewness coefficient was $2.17$), a logarithmic transformation $t_{\log} = \ln(1 + t)$ was performed. No additional filtering or outlier correction was carried out, since all experimental points were obtained under strictly controlled laboratory conditions and are of practical value for analysis, including for studying the boundary regimes of the process. All numerical features used in machine learning models sensitive to scale were standardised using the $Z$-score method: $x_{\text{scaled}} = (x - \mu) / \sigma$, where $\mu$ and $\sigma$ are the mean and standard deviation over the training set.

\subsection{Machine Learning Models}

To solve the multi-task regression problem, 11 algorithms representing the main machine learning paradigms were trained and compared: regularised linear models (Ridge, Lasso, ElasticNet), ensemble methods based on decision trees (Random Forest, Gradient Boosting, XGBoost, CatBoost, Extra Trees), the $k$-nearest neighbours method ($K$ Neighbors), the support vector method (SVR), and a multilayer perceptron (MLP). Simple linear regression was not considered due to feature multicollinearity, and LightGBM was not used in the final comparison, as it did not yield improved results compared with XGBoost and CatBoost. All models were wrapped in \texttt{MultiOutputRegressor}, which ensures independent prediction of each of the four target parameters. The data were split into training ($80\%$, $n = 800$) and test ($20\%$, $n = 200$) sets with stratification by raw material type to preserve the initial distribution of the categorical feature.

\subsection{Hyperparameter Optimisation}

Hyperparameter optimisation was performed using randomised search (\texttt{RandomizedSearchCV}) with 5-fold cross-validation. For each algorithm, a specific range of hyperparameters to be searched was defined: for Random Forest---number of trees $50$--$200$, maximum depth $5$--$15$, minimum number of samples to split a node $2$--$10$; for gradient boosting methods (XGBoost, CatBoost)---number of iterations $50$--$150$, learning rate $0.01$--$0.2$, maximum depth $3$--$6$; for SVR---regularisation parameter $C$ in the range $0.1$--$10$, gamma in the range $0.01$--$0.1$, epsilon $0.01$--$0.2$. The search was conducted with 20 iterations, and the coefficient of determination $R^{2}$ served as the quality criterion.

\subsection{Evaluation Metrics}

The quality of the models was assessed using the following metrics: coefficient of determination
$$
R^{2} = 1 - \frac{\sum (y_{i} - \hat{y}_{i})^{2}}{\sum (y_{i} - \bar{y})^{2}},
$$
root-mean-square error
$$
\text{RMSE} = \sqrt{\frac{1}{n} \sum (y_{i} - \hat{y}_{i})^{2}},
$$
mean absolute error
$$
\text{MAE} = \frac{1}{n} \sum |y_{i} - \hat{y}_{i}|.
$$
Additionally, the normalised root-mean-square error
$$
\text{NRMSE} = \frac{\text{RMSE}}{y_{\max} - y_{\min}}
$$
and the mean absolute percentage error
$$
\text{MAPE} = \frac{100\%}{n} \sum \left| \frac{y_{i} - \hat{y}_{i}}{y_{i}} \right| \quad \text{for} \quad y_{i} \neq 0
$$
were computed. For each target variable separately, the metrics above were computed. The final reported value for each metric is the average over the four target variables.

\subsection{Interpretation and Statistical Analysis}

Interpretation of the obtained results included: calculation of model-independent feature importance by the permutation method (permutation importance), SHAP analysis for tree-based models with visualisation of summary and dependence plots, construction of residual plots and Q--Q plots, as well as a check of the normality of the residual distribution using the Shapiro--Wilk test. Additionally, analysis of variance (ANOVA) was performed to assess the significance of the influence of raw material type on each target variable, and the $t$-test (Mann--Whitney test when normality was violated) was used for pairwise comparison of extraction methods. Differences were considered statistically significant at $p < 0.05$.

\subsection{Software Implementation}

The computations were implemented in Python using the libraries pandas, numpy, scikit-learn, XGBoost, CatBoost, SHAP, LIME, matplotlib, and seaborn. To ensure reproducibility, the random state parameter was fixed (\texttt{random\_state = 42}). The developed preprocessing and prediction pipeline was exported in a production pipeline format with the preservation of all artefacts (models, scalers, encoders, metadata, feature correspondence map). The full code and trained models are available in an open repository.

\section{Formal Problem Statement}
\label{sec:formal-problem-statement}

Let there be an experimental dataset $D = \{(x_{i}, y_{i})\}_{i=1}^{N}$, where $N = 1000$ is the total number of conducted trials. Each vector of input variables $x_{i} \in \mathcal{X} \subset \mathbb{R}^{m}$ characterises the control parameters of the process: $t_{\log} = \ln(1 + t)$ --- the holding time after logarithmic transformation, $T$ --- temperature in degrees Celsius, $P$ --- pressure in atmospheres, $\text{pH}$ --- the acidity of the medium, as well as encoded categorical features (type of raw material and extraction method). The vector of responses $y_{i} \in \mathcal{Y} \subset \mathbb{R}^{k}$ ($k = 4$) includes the key technological quality indicators of the final pectin fraction: pectin yield $Y$ (\%), galacturonic acid content $GA$ (\%), weight-average molecular weight $M_{w}$ (Da), and degree of esterification $DE$ (\%).

The main task consists in constructing a parametric model $f_{\theta}: \mathcal{X} \to \mathcal{Y}$ that provides a reproducible and stable approximation of the mapping $x \mapsto y$ over the entire set of processes that are admissible from a technological standpoint. Model training is treated as the minimisation of the empirical error:
\begin{equation}\label{eq:empirical-risk}
\theta^{*} = \arg\min_{\theta} \sum_{i=1}^{N} \mathcal{L}\bigl( f_{\theta}(x_{i}),\, y_{i} \bigr),
\end{equation}
where the loss function $\mathcal{L}(\cdot, \cdot)$ is chosen as the mean squared error averaged over all target variables:
\begin{equation}\label{eq:loss-function}
\mathcal{L}\bigl( f_{\theta}(x_{i}),\, y_{i} \bigr) = \frac{1}{k} \sum_{j=1}^{k} \bigl( y_{i}^{(j)} - f_{\theta}^{(j)}(x_{i}) \bigr)^{2}.
\end{equation}
Such an approach ensures consistent scaling of errors across quantities of different physical orders (yield is measured in percent, molecular weight in daltons) and allows the criterion to be interpreted as a unified measure of model quality. To improve stability and comparability, the target variables are standardised before training ($Z$-normalisation) according to the formula $y_{\text{scaled}} = (y - \mu_{y}) / \sigma_{y}$, where $\mu_{y}$ and $\sigma_{y}$ are the mean and standard deviation over the training set.

The assessment of accuracy is carried out on the basis of standard regression analysis metrics. The coefficient of determination for each output variable is defined as:
\begin{equation}\label{eq:r2-per-target}
R_{j}^{2} = 1 - \frac{\sum_{i=1}^{n_{\text{test}}} \bigl( y_{i}^{(j)} - \hat{y}_{i}^{(j)} \bigr)^{2}}{\sum_{i=1}^{n_{\text{test}}} \bigl( y_{i}^{(j)} - \bar{y}^{(j)} \bigr)^{2}},
\end{equation}
where $\hat{y}_{i}^{(j)}$ is the model prediction, and $\bar{y}^{(j)}$ is the mean value of the target variable in the test set. The root-mean-square error is computed analogously:
\begin{equation}\label{eq:rmse-per-target}
\text{RMSE}_{j} = \sqrt{\frac{1}{n_{\text{test}}} \sum_{i=1}^{n_{\text{test}}} \bigl( y_{i}^{(j)} - \hat{y}_{i}^{(j)} \bigr)^{2}},
\end{equation}
as well as the mean absolute error:
\begin{equation}\label{eq:mae-per-target}
\text{MAE}_{j} = \frac{1}{n_{\text{test}}} \sum_{i=1}^{n_{\text{test}}} \bigl| y_{i}^{(j)} - \hat{y}_{i}^{(j)} \bigr|.
\end{equation}
The final assessment of model quality is obtained by averaging the corresponding metrics over all $k$ target variables:
\begin{equation}\label{eq:averaged-metrics}
\overline{R^{2}} = \frac{1}{k} \sum_{j=1}^{k} R_{j}^{2}, \qquad
\overline{\text{RMSE}} = \frac{1}{k} \sum_{j=1}^{k} \text{RMSE}_{j}, \qquad
\overline{\text{MAE}} = \frac{1}{k} \sum_{j=1}^{k} \text{MAE}_{j}.
\end{equation}
All metrics are calculated using 5-fold cross-validation, which makes it possible to obtain statistically consistent estimates and construct confidence intervals for the predictions.

The second aspect of the study is related to the problem of optimising the flash hydrolysis regime. It is assumed that the constructed model $f_{\theta^{*}}(x)$ approximates the dependence of the responses on the process parameters sufficiently smoothly, which allows it to be used to search for technological regimes $x$ that provide the desired pectin characteristics. Optimisation is considered as the problem of finding the extremum of the objective function within the set $S$, defined by technological and equipment constraints:
\begin{equation}\label{eq:optimisation}
x^{*} = \arg\max_{x \in S} \; f_{\theta^{*}}^{(\text{yield})}(x),
\end{equation}
where $f_{\theta^{*}}^{(\text{yield})}(x)$ is the model prediction for the pectin yield $Y$. The constraints $S$ are determined by the physical limits of variation of the parameters: $T \in [85, 130]\,^{\circ}\mathrm{C}$ (main regimes $110$--$130\,^{\circ}\mathrm{C}$), $P \in [0.9; 2.2]$~atm, $t \in [3, 10]$~min, $\text{pH} \in [1.5; 2.0]$ (working range $1.8$--$2.0$), as well as the requirement to preserve product quality (e.g., $M_{w} \geq 140$~kDa, $DE \geq 55\%$). In the general case, when simultaneous consideration of several criteria is required (for example, high yield, high molecular weight, and an acceptable degree of esterification), a multi-objective approach is applied: compromise solutions are determined by scalarisation methods or by analysis of the Pareto-optimal set. In the present work, due to the dominant role of pectin yield as the main technological indicator, we restricted ourselves to maximising $Y$ while maintaining $M_{w}$ and $DE$ within physically justified limits set by the technological schedule.

The correctness of the mathematical model and the operability of the optimisation procedures directly depend on the quality of the initial data, correct preprocessing, adequate encoding of categorical features, and the use of stratified cross-validation. Fulfilment of these requirements ensures the reproducibility of the results, conformity with rigorous methodological standards, and the scientific validity of the conclusions drawn.

\section{Experimental Part}
\label{sec:experimental-part}

\subsection{Data Preprocessing and Descriptive Statistics}
\label{subsec:data-preprocessing}

The experimental basis of the study comprised the results of 1,000 independent laboratory trials of flash hydrolysis, carried out under controlled conditions on real process equipment in the Laboratory of High-Molecular Compounds Chemistry of the V.I. Nikitin Institute of Chemistry of the National Academy of Sciences of Tajikistan during the period 2012--2024. Seven types of plant raw material containing pectin were investigated, for each of which the number of experiments ($n$) is given below:
\begin{itemize}
    \item AP(F) (apple pomace of the Faizabad variety) --- $n = 166$;
    \item Qnc. (quince) --- $n = 158$;
    \item AP(M) (apple pomace of the Muminabad variety) --- $n = 156$;
    \item Apr. (apricot) --- $n = 134$;
    \item SFH (sunflower heads) --- $n = 130$;
    \item Rhb. (rhubarb) --- $n = 130$;
    \item Pmp. (pumpkin) --- $n = 126$.
\end{itemize}

\subsubsection{Flash Hydrolysis Methodology and Measured Parameters}

The raw material was pre-ground to a particle size of $0.5$--$1$~mm and normalised by moisture content to $10$--$15\%$ to ensure process reproducibility. Pectin extraction was carried out in an autoclave with the control parameters being varied: temperature in the range $85$--$130\,^{\circ}\mathrm{C}$ (main regimes $110$--$130\,^{\circ}\mathrm{C}$, individual trials at $85\,^{\circ}\mathrm{C}$ for the traditional method), pressure $0.9$--$2.2$~atm, holding time---for the flash method $3$--$10$~min. The acidity of the medium was maintained in the pH interval $1.5$--$2.0$ (working range $1.8$--$2.0$). Upon completion of the holding period, the pressure was instantaneously released, creating a steam-explosion effect that promotes the depolymerisation of protopectin and increases the yield of the target product.

For each trial, four output characteristics of the obtained pectin were determined by analytical methods: pectin yield (PY, \%, gravimetric method); galacturonic acid content (GA, \%, titrimetric method); weight-average molecular weight ($M_{w}$, Da, high-performance liquid chromatography method, HPLC); and degree of esterification (DE, \%, titrimetric method).

\subsubsection{Data Preprocessing}

Primary analysis of the structure of the initial Excel file revealed the presence of 18 columns, of which 12 were auxiliary (Unnamed: 11 -- Unnamed: 17). The columns Unnamed: 11, Unnamed: 12, and Unnamed: 15 were completely empty (100\% missing values), while Unnamed: 13, 14, 16, 17 contained from 98.7\% to 99.5\% missing values (the few non-empty cells were text subheadings of the original Russian-language table). All auxiliary columns were excluded from further analysis.

\begin{figure}[htbp]
    \centering
    \includegraphics[width=0.8\linewidth]{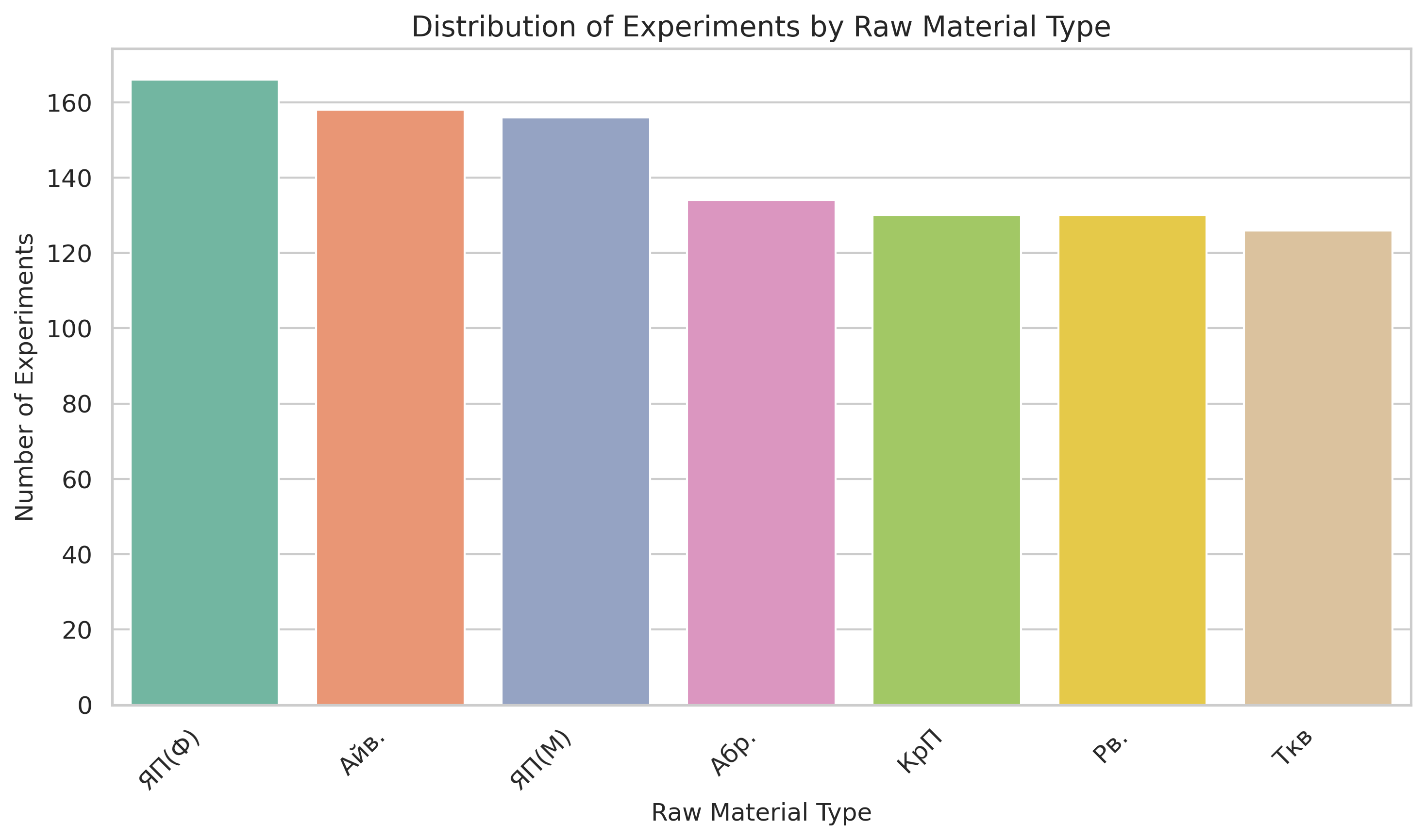}
    \caption{Distribution of the number of experiments across the seven types of plant raw material. The largest number of trials was conducted for apple pomace of the Faizabad variety ($n = 166$), the smallest for pumpkin ($n = 126$).}
    \label{fig:sample-distribution}
\end{figure}

After removing the auxiliary columns, the Russian-language feature names were renamed to their English-language equivalents: \texttt{Obrazets\_pektina} $\to$ \texttt{sample}, \texttt{t\_min} $\to$ \texttt{time\_min}, \texttt{T\_C} $\to$ \texttt{temperature\_c}, \texttt{P\_atm} $\to$ \texttt{pressure\_atm}, \texttt{pH} $\to$ \texttt{ph}, \texttt{T\_Zh} $\to$ \texttt{solid\_liquid\_ratio}, \texttt{PV\_\%} $\to$ \texttt{pectin\_yield}, \texttt{GK\_\%} $\to$ \texttt{galacturonic\_acid}, \texttt{Mw\_D} $\to$ \texttt{molecular\_weight}, \texttt{SE\_\%} $\to$ \texttt{esterification\_degree}. A correspondence map (column mapping) was compiled and saved to ensure full reproducibility of the analysis. The distribution of experiments across the seven investigated types of raw material is clearly presented in Figure~\ref{fig:sample-distribution}, where the numbers of trials for each type of raw material fully correspond to the statistical data given above.

The categorical feature ``type of raw material'' (\texttt{sample}) was transformed by integer encoding (label encoding): Apr. $= 0$, Qnc. $= 1$, SFH $= 2$, Rhb. $= 3$, Pmp. $= 4$, AP(M) $= 5$, AP(F) $= 6$. Additionally, a binary feature \texttt{method\_encoded} was introduced, separating the experiments by extraction method: the value $1$ was assigned to trials with a holding time $\leq 15$~min (flash method), and the value $0$ to trials with a time $> 15$~min (traditional method). According to this criterion, 870 trials (87.0\%) were classified as the flash method, and 130 trials (13.0\%) as the traditional method.

Analysis of the distributions of the numerical variables revealed that the holding time \texttt{time\_min} has a significant positive asymmetry (skewness coefficient $= 2.17$, with a median of $7$~min, a mean of $13.68$~min, and a maximum value of $60$~min). Such high asymmetry is due to the presence of two fundamentally different technological regimes: the short-duration flash method ($3$--$10$~min) and the lengthy traditional method (up to $60$~min). To stabilise the variance and improve the properties of the regression models, a logarithmic transformation was performed: $\texttt{log\_time\_min} = \ln(1 + \texttt{time\_min})$. After the transformation, the distribution of the feature became considerably closer to symmetric, which is favourable for scale-sensitive algorithms.

Missing values in the key numerical variables (\texttt{time\_min}, \texttt{temperature\_c}, \texttt{pressure\_atm}, \texttt{ph}, \texttt{pectin\_yield}, \texttt{galacturonic\_acid}, \texttt{molecular\_weight}, \texttt{esterification\_degree}) were completely absent---all 1,000 rows were 100\% complete. This made it possible to use the full volume of the sample without applying imputation procedures. No additional filtering or winsorisation of outliers was performed, since all experimental points were obtained under strictly controlled laboratory conditions, and the extreme values (e.g., $\texttt{time\_min} = 60$~min) reflect the real boundary regimes of the traditional method and are of practical value.

Summary descriptive statistics of the eight quantitative variables before standardisation are presented in Table~\ref{tab:descriptive-statistics}. For each variable, the following are indicated: arithmetic mean (mean), standard deviation (std), minimum and maximum values (min, max), quartiles $Q_{1}$ (25\%), median (50\%), $Q_{3}$ (75\%), coefficient of variation (CV, \%), and skewness coefficient (skewness).

\begin{table}[htbp]
\centering
\caption{Summary descriptive statistics of input and output variables ($n = 1000$)}
\label{tab:descriptive-statistics}
\begin{tabular}{lccccccccc}
\toprule
\textbf{Variable} & \textbf{mean} & \textbf{std} & \textbf{min} & $\boldsymbol{Q_{1}}$ & \textbf{median} & $\boldsymbol{Q_{3}}$ & \textbf{max} & \textbf{CV, \%} & \textbf{skewness} \\
\midrule
time\_min (min) & 13.68 & 17.98 & 5.0 & 5.0 & 7.0 & 10.0 & 60.0 & 131.4 & 2.17 \\
temperature\_c ($^{\circ}$C) & 118.05 & 13.48 & 85.0 & 120.0 & 120.0 & 130.0 & 130.0 & 11.4 & --1.74 \\
pressure\_atm (atm) & 1.656 & 0.317 & 0.90 & 1.58 & 1.69 & 1.85 & 2.24 & 19.2 & --0.89 \\
pH & 1.931 & 0.126 & 1.50 & 1.85 & 2.00 & 2.00 & 2.08 & 6.5 & --1.47 \\
pectin\_yield (\%) & 20.749 & 4.761 & 12.76 & 14.76 & 21.11 & 25.45 & 29.18 & 22.9 & --0.21 \\
galacturonic\_acid (\%) & 58.828 & 8.958 & 41.70 & 51.05 & 57.75 & 67.55 & 74.42 & 15.2 & --0.07 \\
molecular\_weight (Da) & 136979 & 30088 & 83090 & 105988 & 146902 & 164019 & 190844 & 22.0 & --0.25 \\
esterification\_degree (\%) & 70.292 & 5.826 & 55.24 & 67.11 & 69.53 & 75.17 & 83.04 & 8.3 & --0.06 \\
\bottomrule
\end{tabular}
\end{table}

Analysis of Table~\ref{tab:descriptive-statistics} allows several important conclusions to be drawn. The median conditions of the experiments are: temperature $120\,^{\circ}\mathrm{C}$, pressure $1.69$~atm, holding time $7$~min, pH $2.00$. The galacturonic acid content varies over a wide range---from $41.7$ to $74.4\%$ with a median of $57.8\%$, which reflects the strong influence of the raw material type. The molecular weight varies from $83.1$ to $190.8$~kDa (median $146.9$~kDa), with the interquartile range $Q_{1}$--$Q_{3}$ being $106.0$--$164.0$~kDa. The high concentration of values in this range (50\% of all observations) indicates the stability of the formation of pectin polymer chains under the chosen technological regimes.

The high coefficient of variation for the holding time (CV $= 131.4\%$) is due to the combination of short-duration flash experiments ($3$--$10$~min) and lengthy traditional ones (up to $60$~min). Pectin yield (CV $= 22.9\%$) and molecular weight (CV $= 22.0\%$) are also characterised by significant variability, whereas the degree of esterification is the most stable (CV $= 8.3\%$). The negative values of the skewness coefficient for temperature ($-1.74$), pressure ($-0.89$), and pH ($-1.47$) mean that the ``tail'' of the distribution is stretched to the left, and the bulk of the values is shifted to the right (towards high parameter values).

\begin{figure}[htbp]
    \centering
    \includegraphics[width=0.8\linewidth]{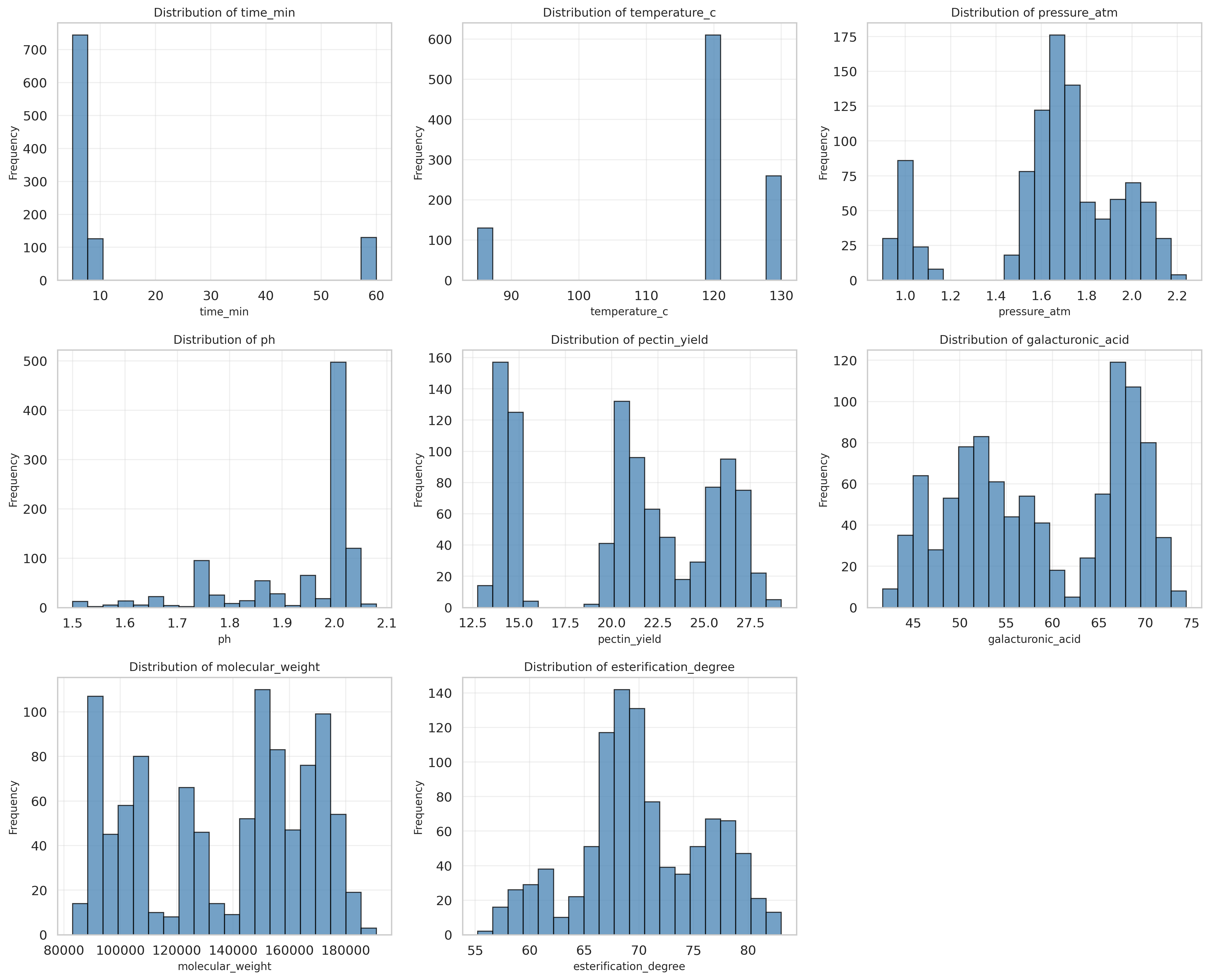}
    \caption{Histograms of the distributions of input technological parameters (upper row) and output product characteristics (lower row). Dashed lines indicate median values.}
    \label{fig:numeric-distributions}
\end{figure}

The histograms of the distributions of all eight numerical variables, presented in Figure~\ref{fig:numeric-distributions}, visualise the key features identified during the descriptive statistical analysis. In the upper row of the figure, reflecting the input technological parameters, the distribution of the holding time (\texttt{time\_min}) confirms the presence of a sharp right-sided asymmetry: the bulk of the observations are concentrated in the interval $3$--$10$~min (flash method), whilst a minor peak is registered in the range $50$--$60$~min, corresponding to traditional extraction. The distribution of temperature (\texttt{temperature\_c}) exhibits a bimodal character with two pronounced modes---near $85$--$90\,^{\circ}\mathrm{C}$ and in the interval $120$--$130\,^{\circ}\mathrm{C}$, which is due to differences in the temperature regimes applied for different types of raw material and experimental procedures. The pressure variable (\texttt{pressure\_atm}) is characterised by a distribution close to normal, with a centre around $1.6$--$1.8$~atm and a slight ``tail'' towards low values (around $1.0$~atm). The pH values, in turn, are concentrated in the narrow range $1.9$--$2.0$, which reflects the maintenance of a stable acidic medium throughout the entire hydrolysis process.

In the lower row of Figure~\ref{fig:numeric-distributions}, illustrating the output characteristics of the product, the distribution of pectin yield (\texttt{pectin\_yield}) has a clearly pronounced bimodal structure with peaks around $14$--$15\%$ and $25$--$27\%$, which testifies to the significant dependence of this indicator on the type of raw material and processing conditions. The galacturonic acid content (\texttt{galacturonic\_acid}) is distributed more uniformly; however, a weak right-sided asymmetry is observed, with the main peak in the range $65$--$70\%$. The distribution of molecular weight (\texttt{molecular\_weight}) is of a diffuse multimodal character, with value intervals from $80$ to $190$~kDa, reflecting the diversity of polymer structures of the obtained pectins. The degree of esterification (\texttt{esterification\_degree}) is close to a normal distribution, with a median around $69.5\%$ and a high concentration of values in the interval $65$--$75\%$. On the whole, Figure~\ref{fig:numeric-distributions} clearly demonstrates the complex nature of the investigated sample, combining both the technological features of the process (bimodality of temperature and holding time) and a wide spectrum of properties of the obtained pectins, which must be taken into account in the subsequent construction of predictive models.

For algorithms sensitive to feature scale (SVR, MLP, $K$ Neighbors, regularised linear models), $Z$-standardisation of all continuous variables was performed: $x_{\text{scaled}} = (x - \mu) / \sigma$, where $\mu$ and $\sigma$ are the mean and standard deviation calculated over the training set ($n = 800$). Categorical features (\texttt{sample\_encoded}, \texttt{method\_encoded}) were not scaled. The final feature matrix for machine learning included 6 variables: \texttt{log\_time\_min} (logarithm of holding time), \texttt{temperature\_c} (temperature), \texttt{pressure\_atm} (pressure), \texttt{ph} (acidity), \texttt{sample\_encoded} (type of raw material), \texttt{method\_encoded} (extraction method). The target block comprised 4 variables: \texttt{pectin\_yield}, \texttt{galacturonic\_acid}, \texttt{molecular\_weight}, \texttt{esterification\_degree}.

Thus, after a complete preprocessing cycle, a clean, fully complete dataset of dimensionality $1000 \times 10$ (6 features + 4 targets) was formed, containing no missing values and ready for the modelling stage. All preprocessing procedures were recorded in log files; intermediate results (descriptive statistics, column correspondence map, information on data types and missing values) were exported to the \texttt{results/} directory to ensure full reproducibility.

\subsection{Correlation Analysis and Analysis of Variance (ANOVA)}
\label{subsec:correlation-anova}

\begin{figure}[htbp]
    \centering
    \includegraphics[width=0.8\linewidth]{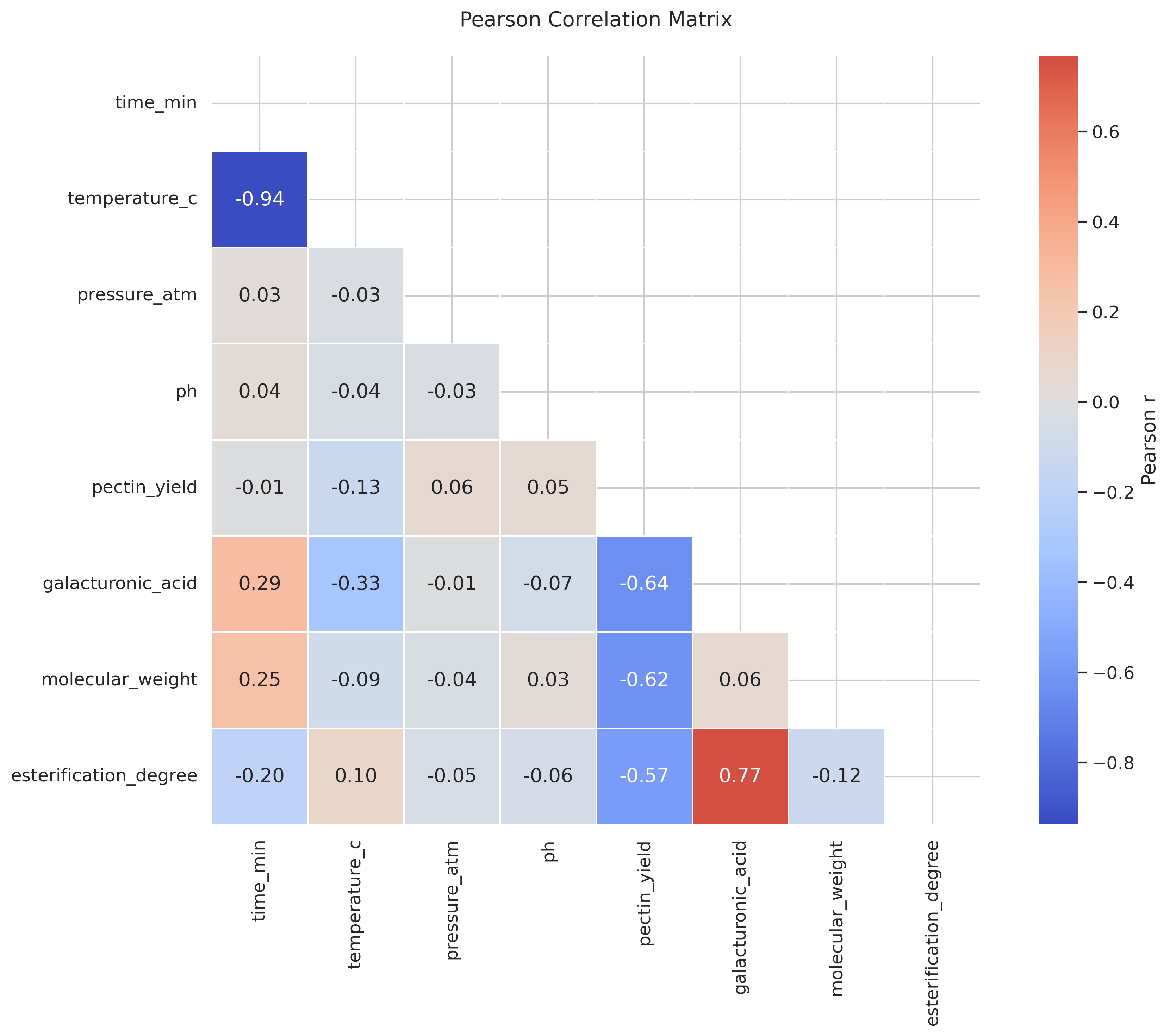}
    \caption{Pearson correlation heat map.}
    \label{fig:correlation-matrix}
\end{figure}

To reveal the structure of linear relationships between the technological parameters and the output characteristics of the flash hydrolysis process, a comprehensive correlation analysis was carried out. The full matrix of pairwise Pearson linear correlation coefficients ($r$) was calculated for eight quantitative variables. The statistical significance of each coefficient was assessed using a two-tailed $t$-test; the null hypothesis $H_{0}: r = 0$ was rejected at $p < 0.05$. The full results, indicating the correlation coefficients and the corresponding $p$-values, are summarised in Table~\ref{tab:correlation-matrix}. A heat map with annotation of the $r$ values was constructed for a clear presentation of the results (Figure~\ref{fig:correlation-matrix}).

Analysis of the presented correlation matrix (Table~\ref{tab:correlation-matrix}, Figure~\ref{fig:correlation-matrix}) allows the following key regularities to be identified. The strongest negative correlation is observed between the holding time (\texttt{time\_min}) and temperature (\texttt{temperature\_c}): $r = -0.94$ ($p < 0.001$). This relationship is of an expected nature and is explained by the fact that, within the framework of the experimental design, higher temperatures ($120$--$130\,^{\circ}\mathrm{C}$) were applied predominantly during short-duration flash hydrolysis ($3$--$10$~min), whereas traditional extraction at lower temperatures (around $90\,^{\circ}\mathrm{C}$) required a lengthy holding period (up to $60$~min). Such a high inverse correlation testifies to the close interrelation of these two control parameters, which must be taken into account when constructing regression models.

Significant correlations were also revealed between the output characteristics of the product. The galacturonic acid content (\texttt{galacturonic\_acid}) demonstrates a pronounced positive correlation with the degree of esterification (\texttt{esterification\_degree}): $r = 0.77$ ($p < 0.001$). This indicates that, in the investigated pectin samples, an increase in galacturonic acid content is, as a rule, accompanied by an increase in the degree of esterification. At the same time, both of these variables have negative correlation relationships with the pectin yield (\texttt{pectin\_yield}): $r = -0.64$ ($p < 0.001$) for galacturonic acid and $r = -0.57$ ($p < 0.001$) for the degree of esterification. This fact suggests the existence of a trade-off relationship: samples with a higher pectin yield are, as a rule, characterised by a lower galacturonic acid content and a lower degree of esterification.

The molecular weight (\texttt{molecular\_weight}) also has a negative correlation with the pectin yield ($r = -0.62$; $p < 0.001$), which is consistent with the observation of a decrease in the molecular weight of pectin with an increase in yield, probably due to more profound depolymerisation of protopectin during the hydrolysis process. In turn, the holding time (\texttt{time\_min}) correlates positively with the galacturonic acid content ($r = 0.29$; $p < 0.001$) and with the molecular weight ($r = 0.25$; $p < 0.001$), whereas the relationship with pectin yield is practically absent ($r = -0.01$; $p = 0.822$). The remaining correlation relationships between the technological parameters and the output characteristics lie in the range of weak values ($|r| < 0.15$) or are statistically insignificant ($p > 0.05$), which testifies to the complex non-linear nature of the process under investigation. For a more detailed analysis of the influence of the type of raw material and processing conditions on the properties of pectin, an analysis of variance (ANOVA) was carried out, the results of which are presented in the following subsection.

\begin{table}[htbp]
\centering
\caption{Pearson correlation matrix (lower triangle --- $r$, upper triangle --- $p$-value)}
\label{tab:correlation-matrix}
\resizebox{\textwidth}{!}{%
\begin{tabular}{lcccccccc}
\toprule
\textbf{Variable} & \textbf{time\_min} & \textbf{temp\_c} & \textbf{press\_atm} & \textbf{pH} & \textbf{pectin\_y} & \textbf{gal\_acid} & \textbf{mol\_wt} & \textbf{ester\_deg} \\
\midrule
time\_min & 1.000 & $<0.001$ & 0.430 & 0.261 & 0.822 & $<0.001$ & $<0.001$ & $<0.001$ \\
temperature\_c & $-0.938$ & 1.000 & 0.343 & 0.190 & $<0.001$ & $<0.001$ & 0.006 & 0.002 \\
pressure\_atm & 0.025 & $-0.030$ & 1.000 & 0.351 & 0.074 & 0.715 & 0.216 & 0.146 \\
pH & 0.036 & $-0.042$ & $-0.030$ & 1.000 & 0.083 & 0.020 & 0.359 & 0.049 \\
pectin\_yield & $-0.007$ & $-0.128$ & 0.056 & 0.055 & 1.000 & $<0.001$ & $<0.001$ & $<0.001$ \\
galacturonic\_acid & 0.287 & $-0.329$ & $-0.012$ & $-0.073$ & $-0.639$ & 1.000 & 0.046 & $<0.001$ \\
molecular\_weight & 0.251 & $-0.087$ & $-0.039$ & 0.029 & $-0.621$ & 0.063 & 1.000 & $<0.001$ \\
esterification\_degree & $-0.195$ & 0.099 & $-0.046$ & $-0.062$ & $-0.574$ & 0.768 & $-0.124$ & 1.000 \\
\bottomrule
\end{tabular}%
}
\end{table}

\textbf{Relationships between input parameters.} The strongest correlation in the entire matrix was found between the holding time and temperature: $r = -0.938$ ($p < 0.001$). This virtually functional inverse dependence reflects an objective technological regularity: lengthy processes (traditional method, up to $60$~min) are conducted at reduced temperatures ($85$--$110\,^{\circ}\mathrm{C}$), whereas short-duration high-temperature treatment (flash method, $3$--$10$~min) is realised at $120$--$130\,^{\circ}\mathrm{C}$. Such high multicollinearity ($|r| > 0.9$) explains the low efficiency of linear models and justifies the expediency of the logarithmic transformation of time. Pressure and pH demonstrate weak, statistically insignificant correlations both between themselves ($r = -0.030$, $p = 0.351$) and with other input parameters (all $|r| < 0.06$, $p > 0.05$), which makes them independent control factors, albeit with a limited influence on the output characteristics.

\textbf{Relationships between output characteristics.} All four target variables correlate statistically significantly with one another ($p < 0.001$ for six pairs out of six). The closest positive relationship was found between the galacturonic acid content and the degree of esterification: $r = 0.768$ ($p < 0.001$). This indicates the coupling of the de-esterification processes and the breakdown of the polygalacturonic chain: the preservation of methoxyl groups (high DE) is associated with a high GA content in the product. The pectin yield (\texttt{pectin\_yield}) correlates negatively with all three quality indicators: $r = -0.639$ with GA, $r = -0.621$ with $M_{w}$, $r = -0.574$ with DE (all $p < 0.001$). Thus, within the investigated parameter range, the maximum product yield is often achieved at the cost of a reduction in its quality---lower molecular weight, lower galacturonic acid content, and a reduced degree of esterification. This ``yield vs.\ quality'' dilemma is a classic one for chemical technology and requires multi-criteria optimisation, which justifies the necessity of applying multi-task machine learning methods.

\textbf{Input--output relationships.} Among the technological parameters, temperature exerts the strongest influence on the output characteristics: it correlates negatively with the GA content ($r = -0.329$, $p < 0.001$) and, more weakly, with the pectin yield ($r = -0.128$, $p < 0.001$). The holding time (on the original scale) is positively related to the GA content ($r = 0.287$, $p < 0.001$) and molecular weight ($r = 0.251$, $p < 0.001$), but negatively to the degree of esterification ($r = -0.195$, $p < 0.001$). These dependences are fully consistent with the known physicochemical mechanisms of the acid hydrolysis of protopectin: an increase in temperature accelerates the degradation of the polygalacturonic chain (decrease in GA and $M_{w}$), whereas an increase in duration promotes a more complete extraction of pectin, but simultaneously intensifies de-esterification. Pressure did not demonstrate statistically significant correlations with the output variables (all $|r| < 0.08$, $p > 0.05$). For pH, a weak but statistically significant negative correlation with the galacturonic acid content ($r = -0.073$, $p = 0.020$) was found, which may be related to random variation with a large sample size; the remaining correlations of pH are insignificant ($p > 0.05$).

\begin{figure}[htbp]
    \centering
    \includegraphics[width=0.8\linewidth]{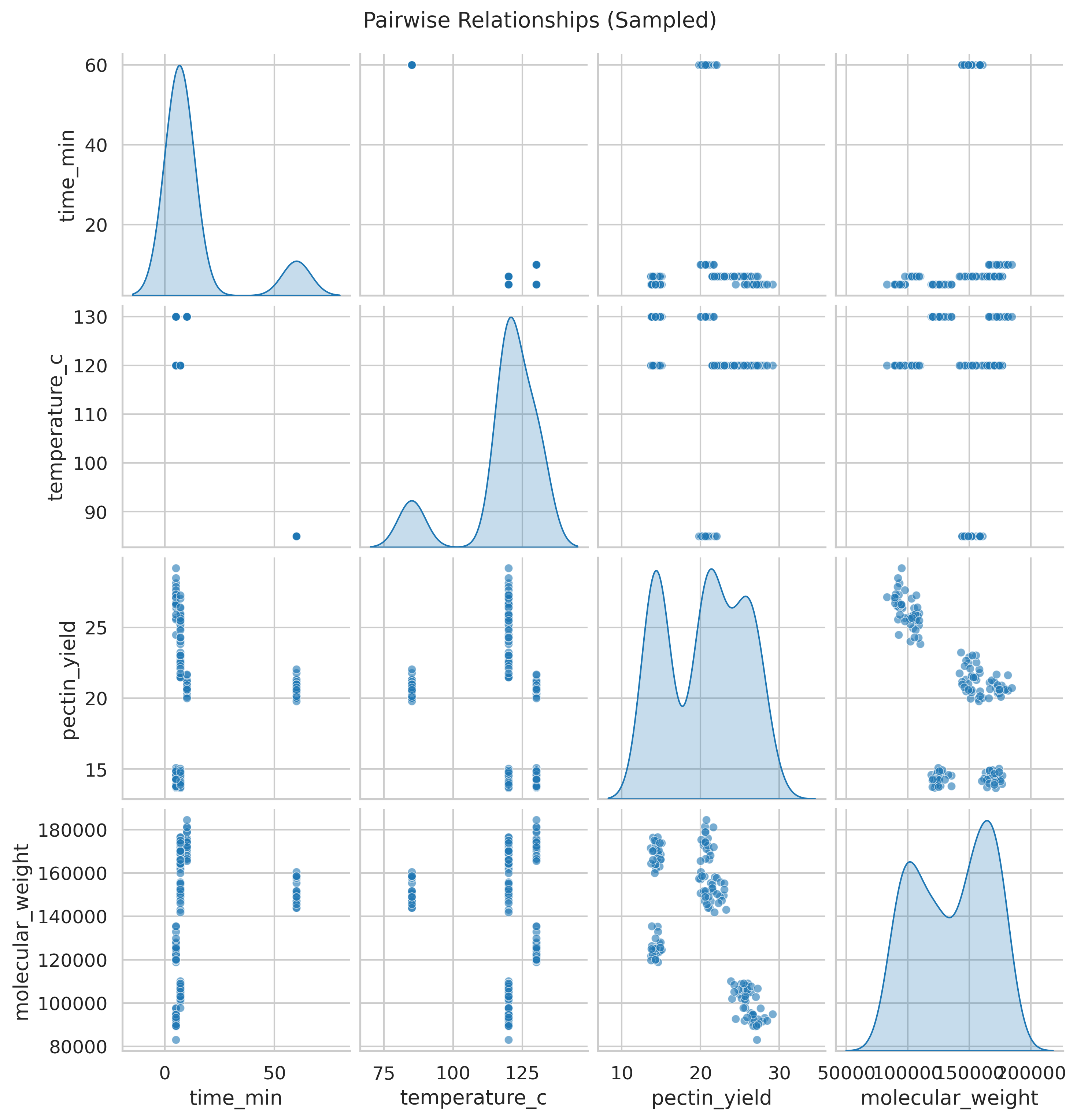}
    \caption{Matrix of pairwise scatter plots (pairplot). On the diagonal --- kernel density estimates (KDE), below the diagonal --- scatter plots, above the diagonal --- 95\% scatter ellipses.}
    \label{fig:pairwise-relationships}
\end{figure}

\textbf{Visualisation of pairwise relationships.} For visualising the pairwise relationships and the cluster structure of the data, a matrix of scatter plots was constructed with points coloured by raw material type (pairplot), presented in Figure~\ref{fig:pairwise-relationships}. On the diagonal of the matrix are the kernel density estimates (KDE), below the diagonal are the scatter plots, and above the diagonal are the scatter ellipses corresponding to the 95\% confidence region for the multivariate normal distribution of each raw material type. Colouring was performed according to the seven types of plant raw material.

Analysis of Figure~\ref{fig:pairwise-relationships} revealed a clear clustering of the data by raw material type, confirming the dominant role of the chemical composition of the initial material in shaping the properties of the final product. Apple pomace of both varieties (AP(F) and AP(M)) are concentrated in the region of high molecular weight values ($140$--$180$~kDa) and high galacturonic acid content ($65$--$75\%$), which is explained by the high degree of polymerisation of the native pectin complexes of apples. Apricot and rhubarb, on the contrary, form compact clusters in the zone of low molecular weights ($80$--$120$~kDa) and reduced galacturonic acid content ($45$--$55\%$). The scatter ellipses confirm the specificity of the technological regimes for different raw material types. The absence of significant overlap of the ellipses for different raw material types justifies the need for stratification of the sample according to this feature when splitting into training and test sets.

\textbf{Analysis of variance (ANOVA).} For a formal statistical assessment of the influence of the categorical feature ``type of raw material'' on each of the four target variables, a one-way analysis of variance (one-way ANOVA) was performed. The null hypothesis $H_{0}$ was tested: the mean values of the target variable are equal for all seven raw material types. The results are presented in Table~\ref{tab:anova-results}.

\begin{table}[htbp]
\centering
\caption{Results of one-way analysis of variance (ANOVA): influence of raw material type on the output characteristics of pectin}
\label{tab:anova-results}
\begin{tabular}{lccc}
\toprule
\textbf{Target variable} & $\boldsymbol{F}$\textbf{-statistic} & $\boldsymbol{p}$\textbf{-value} & \textbf{Significance} ($\alpha = 0.05$) \\
\midrule
pectin\_yield & 9292.95 & $< 0.001$ & Yes \\
galacturonic\_acid & 3929.38 & $< 0.001$ & Yes \\
molecular\_weight & 8182.43 & $< 0.001$ & Yes \\
esterification\_degree & 1183.27 & $< 0.001$ & Yes \\
\bottomrule
\end{tabular}
\end{table}

For all four target variables, the null hypothesis of equality of the group means is rejected at an exceptionally high level of significance ($p < 0.001$). The extremely high values of the $F$-statistic (from $1183$ to $9293$) testify that the between-group variance, caused by differences in raw material types, manifoldly exceeds the random within-group variation. The type of raw material most strongly differentiates the pectin yield ($F = 9293$) and the molecular weight ($F = 8182$). These results serve as a rigorous statistical justification for including \texttt{sample\_encoded} among the predictors and predict its dominant role in the machine learning models.

\begin{figure}[htbp]
    \centering
    \includegraphics[width=0.8\linewidth]{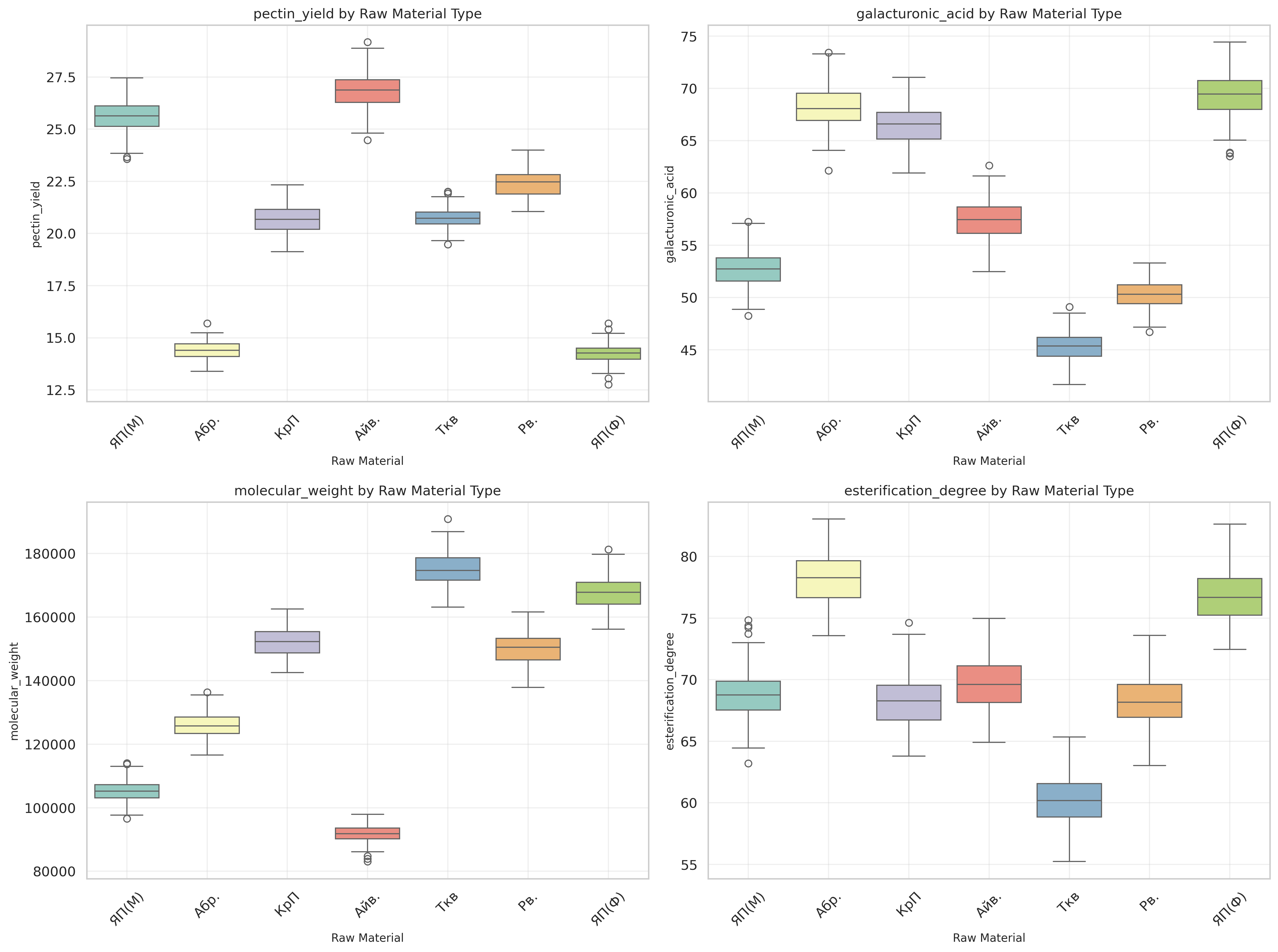}
    \caption{Distribution of the output characteristics of pectin (box plots) across the seven types of plant raw material.}
    \label{fig:targets-by-sample}
\end{figure}

Figure~\ref{fig:targets-by-sample} presents box plots illustrating the distribution of the output characteristics of pectin across the seven types of plant raw material. As can be seen from the figure, quince and apple pomace of the Muminabad variety are characterised by the highest values of pectin yield with a compact interquartile range, whereas apricot and apple pomace of the Faizabad variety show low yields. Similar regularities are observed for the other indicators as well: for example, apple pomace (AP(F) and AP(M)) are concentrated in the region of high molecular weight values and high galacturonic acid content, whilst apricot and pumpkin form clusters with reduced values of these parameters.

Thus, the correlation and analysis of variance revealed the following key regularities that determine the machine learning strategy: (1) strong multicollinearity of time and temperature ($r = -0.94$), requiring the use of ensemble methods robust to correlations; (2) a negative correlation between the yield and the quality indicators of pectin ($|r| \approx 0.6$), justifying the multi-task approach; (3) the overwhelming influence of the raw material type on all target variables (ANOVA, $p < 0.001$), predetermining its role as a key predictor. The conclusions drawn are fully consistent with the physicochemical understanding of the flash hydrolysis mechanism and create a reliable foundation for the stage of constructing predictive models.

\subsection{Comparison of Machine Learning Model Performance}
\label{subsec:model-comparison}

To solve the problem of multi-task prediction of the four output characteristics of pectin, eleven machine learning algorithms covering the main methodological paradigms were developed, trained, and compared:
\begin{itemize}
    \item \textbf{Regularised linear models:} Ridge Regression (L2-regularisation, final parameter $\alpha = 1.0$), Lasso Regression (L1-regularisation, $\alpha = 0.1$, \texttt{max\_iter} = 10000), ElasticNet (combination of L1 and L2, $\alpha = 0.01$, \texttt{l1\_ratio} = 0.5, \texttt{max\_iter} = 10000). Simple linear regression without regularisation was excluded from consideration due to its complete lack of informativeness under conditions of multicollinearity ($r = -0.94$ between time and temperature) and the absence of tunable hyperparameters.
    \item \textbf{Ensemble methods based on decision trees:} Random Forest (\texttt{n\_estimators} = 100, \texttt{max\_depth} = 10, \texttt{min\_samples\_split} = 5), Gradient Boosting (\texttt{n\_estimators} = 100, \texttt{learning\_rate} = 0.1, \texttt{max\_depth} = 4, \texttt{subsample} = 0.8), XGBoost (\texttt{n\_estimators} = 100, \texttt{learning\_rate} = 0.1, \texttt{max\_depth} = 4), CatBoost (\texttt{iterations} = 100, \texttt{learning\_rate} = 0.1, \texttt{depth} = 4), Extra Trees (\texttt{n\_estimators} = 100, \texttt{max\_depth} = 10).
    \item \textbf{Kernel methods and distance-based methods:} SVR (RBF kernel, $C = 1.0$, \texttt{gamma} = `scale', $\varepsilon = 0.1$), $K$-Neighbors ($k = 5$, \texttt{weights} = `distance').
    \item \textbf{Neural networks:} MLP (two hidden layers: 100 and 50 neurons, ReLU activation function, \texttt{max\_iter} = 1000).
\end{itemize}

All models were implemented in the \texttt{MultiOutputRegressor} architecture of the scikit-learn library, which trains an independent regressor for each of the four target variables using a common set of six input features. The data were split into training ($80\%$, $n = 800$) and test ($20\%$, $n = 200$) sets with stratification by the feature \texttt{sample\_encoded} to preserve the initial distribution of raw material types in both sets. All continuous features were standardised using the $Z$-score method; categorical features (\texttt{sample\_encoded}, \texttt{method\_encoded}) were left on the original integer scale.

For each algorithm, hyperparameter optimisation was performed using randomised search (\texttt{RandomizedSearchCV}) with 5-fold cross-validation. The number of search iterations was 20, and the quality criterion was the coefficient of determination $R^{2}$. Detailed grids of the hyperparameters searched for each algorithm are given in the Materials and Methods section. To ensure full reproducibility of the results, the random state parameter was fixed at \texttt{random\_state = 42}.

\begin{figure}[htbp]
    \centering
    \includegraphics[width=0.8\linewidth]{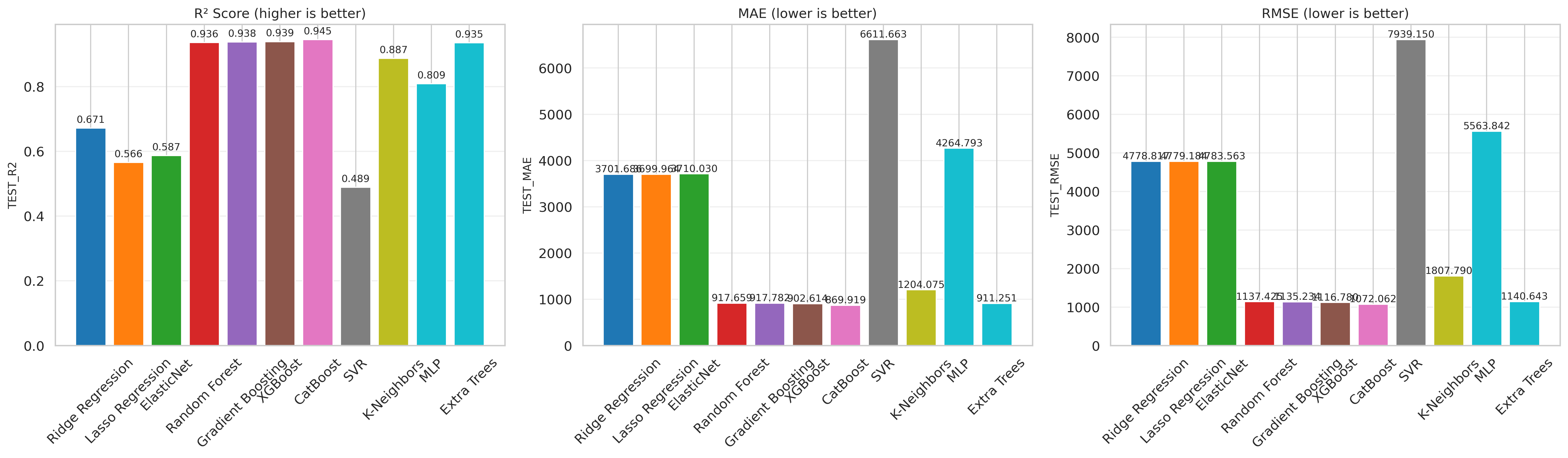}
    \caption{Comparison of the performance of eleven machine learning algorithms on the test set.}
    \label{fig:model-comparison}
\end{figure}

The comparative results of the eleven models on the held-out test set ($n = 200$) are illustrated in Figure~\ref{fig:model-comparison}, where the metrics $R^{2}$, MAE, and RMSE, averaged over the four target variables, are presented for each model. The numerical values of the corresponding metrics are given in Table~\ref{tab:model-comparison}.

\begin{table}[htbp]
\centering
\caption{Comparison of machine learning model performance (metrics averaged over 4 target variables, test set)}
\label{tab:model-comparison}
\begin{tabular}{lcccc}
\toprule
\textbf{Model} & \textbf{Avg Test $R^{2}$} & \textbf{Avg Test MAE} & \textbf{Avg Test RMSE} & \textbf{Training time, s} \\
\midrule
Ridge Regression & 0.6714 & 3701.69 & 4778.82 & $\approx 0.1$ \\
Lasso Regression & 0.5658 & 3699.96 & 4779.18 & $\approx 0.1$ \\
ElasticNet & 0.5865 & 3710.03 & 4783.56 & $\approx 0.1$ \\
Random Forest & 0.9364 & 917.66 & 1137.42 & 0.9 \\
Gradient Boosting & 0.9378 & 917.78 & 1135.23 & 0.4 \\
XGBoost & 0.9389 & 902.61 & 1116.78 & 0.2 \\
CatBoost & 0.9449 & 869.92 & 1072.06 & 0.2 \\
SVR & 0.4888 & 6611.66 & 7939.15 & 0.2 \\
$K$-Neighbors & 0.8870 & 1204.08 & 1807.79 & $\approx 0.1$ \\
MLP & 0.8087 & 4264.79 & 5563.84 & 26.7 \\
Extra Trees & 0.9353 & 911.25 & 1140.64 & 1.0 \\
\bottomrule
\end{tabular}
\end{table}

Analysis of Figure~\ref{fig:model-comparison} and Table~\ref{tab:model-comparison} allows the following key conclusions to be formulated.

\textbf{Superiority of ensemble methods.} The group of five ensemble algorithms based on decision trees (Random Forest, Gradient Boosting, XGBoost, CatBoost, Extra Trees) demonstrates overwhelming superiority over all other classes of models across all three metrics. The average $R^{2}$ for this group lies in the narrow range $0.935$--$0.945$, whereas the best of the linear models (Ridge) achieves only $0.671$. The gap in prediction quality between the ensembles and the linear models amounts to $\Delta R^{2} \approx 0.27$, which is a very substantial magnitude. This confirms the presence of significant non-linear dependences and interactions between features, which are effectively captured by decision trees but fundamentally inaccessible to linear methods. In addition, the high accuracy of the ensembles in the presence of multicollinearity ($r = -0.94$ between time and temperature) confirms their robustness to correlated predictors.

\textbf{Leader --- CatBoost.} The best result was shown by the CatBoost algorithm: average $R^{2} = 0.9449$, MAE $= 869.92$, RMSE $= 1072.06$. Its advantage over the nearest competitors---XGBoost ($R^{2} = 0.9389$) and Random Forest ($R^{2} = 0.9364$)---is small in absolute terms ($\Delta R^{2} \approx 0.006$), but is stably reproduced across all four target variables and is confirmed by the lowest values of MAE and RMSE. The probable reason for CatBoost's leadership is its built-in mechanism for processing categorical features based on ordered boosting and target variable statistics, which for the feature \texttt{sample\_encoded} (7 gradations, shown by ANOVA to be exceptionally informative) provides an additional gain compared with methods that require preliminary numerical encoding of categories.

\textbf{Low efficiency of SVR, MLP, and linear models.} The support vector method (SVR) showed the worst result among all the tested algorithms: $R^{2} = 0.489$, MAE $= 6611.66$, RMSE $= 7939.15$. This is explained, firstly, by the sample size (1,000 observations) being insufficient for the full realisation of the potential of the RBF kernel, and, secondly, by the presence of a discrete categorical feature with seven gradations, for which kernel methods have no natural way of accounting. The multilayer perceptron (MLP), with a considerably longer training time (26.7~s against 0.2--1.0~s for the ensembles), achieved an $R^{2}$ of only $0.809$; moreover, analysis of the training log showed that the model exhausted the limit of 1,000 iterations without achieving full convergence. The regularised linear models (Ridge, Lasso, ElasticNet) predictably showed low results ($R^{2}$ from $0.566$ to $0.671$), which is to be expected in the presence of multicollinearity and non-linear relationships.

\textbf{Hyperparameter optimisation of CatBoost.} For the best model identified during the comparative analysis (CatBoost), a full cycle of hyperparameter optimisation was carried out. A randomised search over 20 combinations with 5-fold cross-validation made it possible to determine the optimal configuration: \texttt{learning\_rate} = 0.05 (moderate learning rate), \texttt{depth} = 3 (shallow trees to prevent overfitting), \texttt{iterations} = 150 (increased number of boosting iterations), \texttt{l2\_leaf\_reg} = 3 (moderate L2-regularisation on the leaves). The cross-validation $R^{2}$ with the optimal hyperparameters was $0.9478$, and on the independent test set it was $0.9460$. The closeness of these values (a difference of less than $0.002$) testifies to the absence of overfitting and to the good generalisation ability of the model.

The detailed quality metrics of the optimised CatBoost model for each of the four target variables are given in Table~\ref{tab:catboost-metrics}. In addition to the main metrics ($R^{2}$, MAE, RMSE), the following were also computed and are presented: mean absolute percentage error (MAPE), normalised root-mean-square error (NRMSE), and explained variance.

\begin{table}[htbp]
\centering
\caption{Quality metrics of the CatBoost model (after optimisation) for individual target variables}
\label{tab:catboost-metrics}
\begin{tabular}{lccccc}
\toprule
\textbf{Target variable} & $\boldsymbol{R^{2}}$ & \textbf{MAE} & \textbf{MAPE, \%} & \textbf{NRMSE} & \textbf{Explained Variance} \\
\midrule
pectin\_yield & 0.9840 & 0.472 \% & 2.31 & 0.0401 & 0.9840 \\
galacturonic\_acid & 0.9485 & 1.553 \% & 2.64 & 0.0632 & 0.9486 \\
molecular\_weight & 0.9798 & 3476 Da & 2.57 & 0.0413 & 0.9799 \\
esterification\_degree & 0.8673 & 1.665 \% & 2.38 & 0.0793 & 0.8697 \\
\midrule
Average & 0.9449 & --- & 2.48 & 0.056 & 0.9456 \\
\bottomrule
\end{tabular}
\end{table}

Analysis of Table~\ref{tab:catboost-metrics} shows that the pectin yield ($R^{2} = 0.9840$) and the molecular weight ($R^{2} = 0.9798$) are predicted most accurately. For these two critically important indicators, the NRMSE values do not exceed $0.042$, which corresponds to excellent model quality according to criteria generally accepted in engineering applications (NRMSE $< 0.1$ --- high accuracy). The galacturonic acid content is predicted with $R^{2} = 0.9485$ and MAPE $= 2.64\%$, which is also a good result.

The lowest accuracy was achieved for the degree of esterification: $R^{2} = 0.8673$ with MAPE $= 2.38\%$. A certain discrepancy between $R^{2}$ and MAPE for this target is noteworthy: the relatively low $R^{2}$ (0.867) with a low MAPE (2.38\%) testifies that the model makes small systematic errors on a few observations, which are ``penalised'' by the quadratic metric $R^{2}$ more strongly than by the linear MAPE. The NRMSE for DE is $0.079$, which still falls within the high accuracy range ($< 0.1$), but indicates potential for further improvement---possibly by including additional predictors, such as the exact acid concentration or the heating rate. It is important to note that for all four target variables, the MAPE values do not exceed $3\%$, which is substantially lower than the $5\%$ threshold usually adopted in technological control tasks. This makes it possible to recommend the constructed model for practical use in product certification tasks and in the ongoing monitoring of the production process.

\begin{figure}[htbp]
    \centering
    \includegraphics[width=0.8\linewidth]{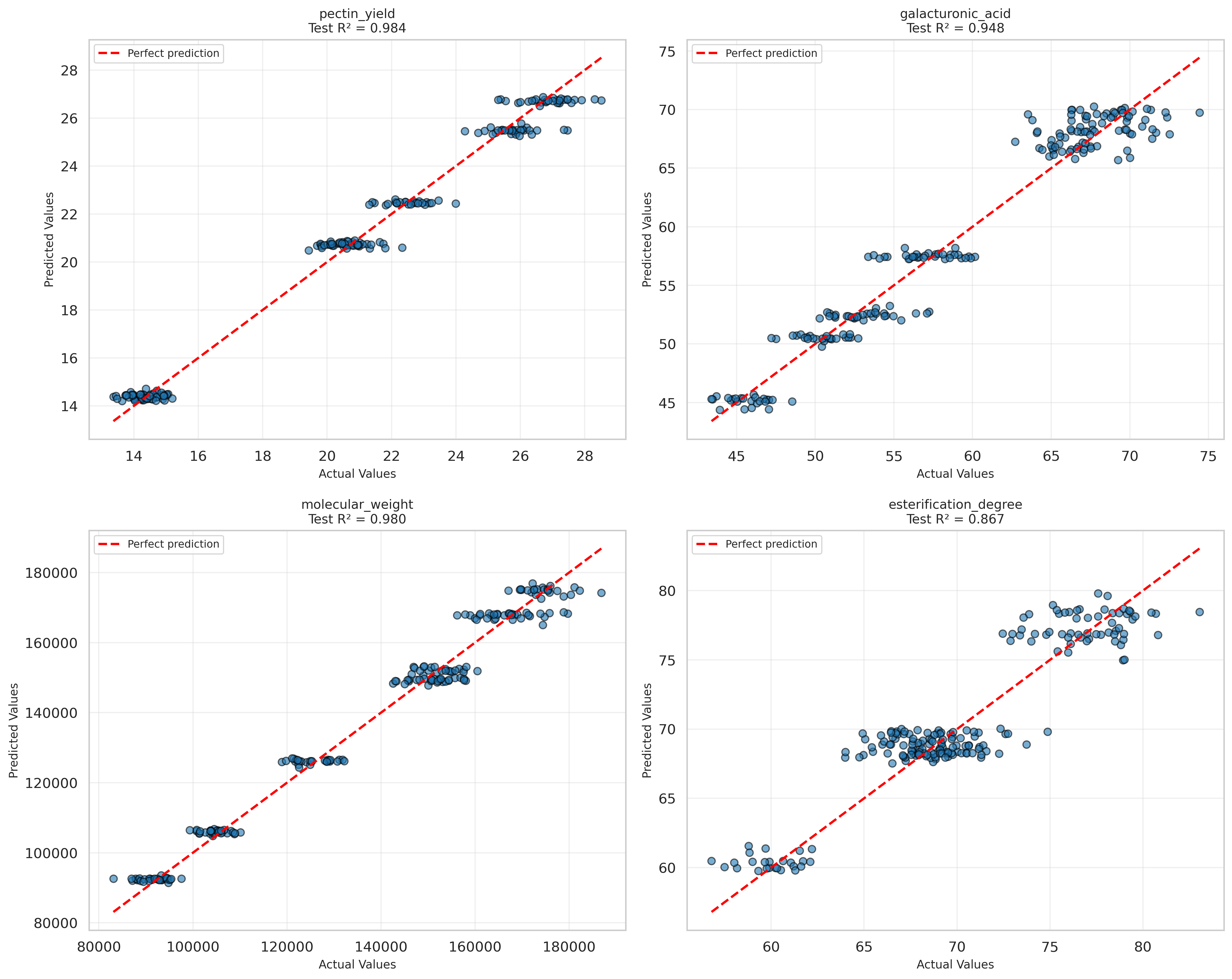}
    \caption{Scatter plots ``Predicted vs.\ Actual Values'' for the CatBoost model across the four target variables. The red dashed line corresponds to the ideal prediction ($y = x$).}
    \label{fig:predictions-vs-actual}
\end{figure}

\textbf{Visualisation of prediction quality.} For a clear assessment of the model's accuracy, scatter plots of ``predicted vs.\ actual values'' were constructed, presented in Figure~\ref{fig:predictions-vs-actual}. These plots make it possible to evaluate not only the numerical metrics but also the nature of the model's errors.

As can be seen from Figure~\ref{fig:predictions-vs-actual}, for all four target variables the points are tightly concentrated along the diagonal line ($y = x$), corresponding to the ideal prediction. Pronounced systematic deviations are absent. The highest density of points near the diagonal is observed for pectin yield and molecular weight, which is fully consistent with the high values of $R^{2}$ (0.984 and 0.980). For the degree of esterification, the scatter of the points is somewhat greater; however, they still cluster around the line of ideal prediction, confirming the statistically acceptable quality of the model even for the least accurately predicted variable. Thus, Figure~\ref{fig:predictions-vs-actual} clearly confirms the conclusions drawn from the analysis of the numerical metrics and demonstrates the high reliability of the constructed model for all the investigated output parameters of pectin.

\begin{figure}[htbp]
    \centering
    \includegraphics[width=0.8\linewidth]{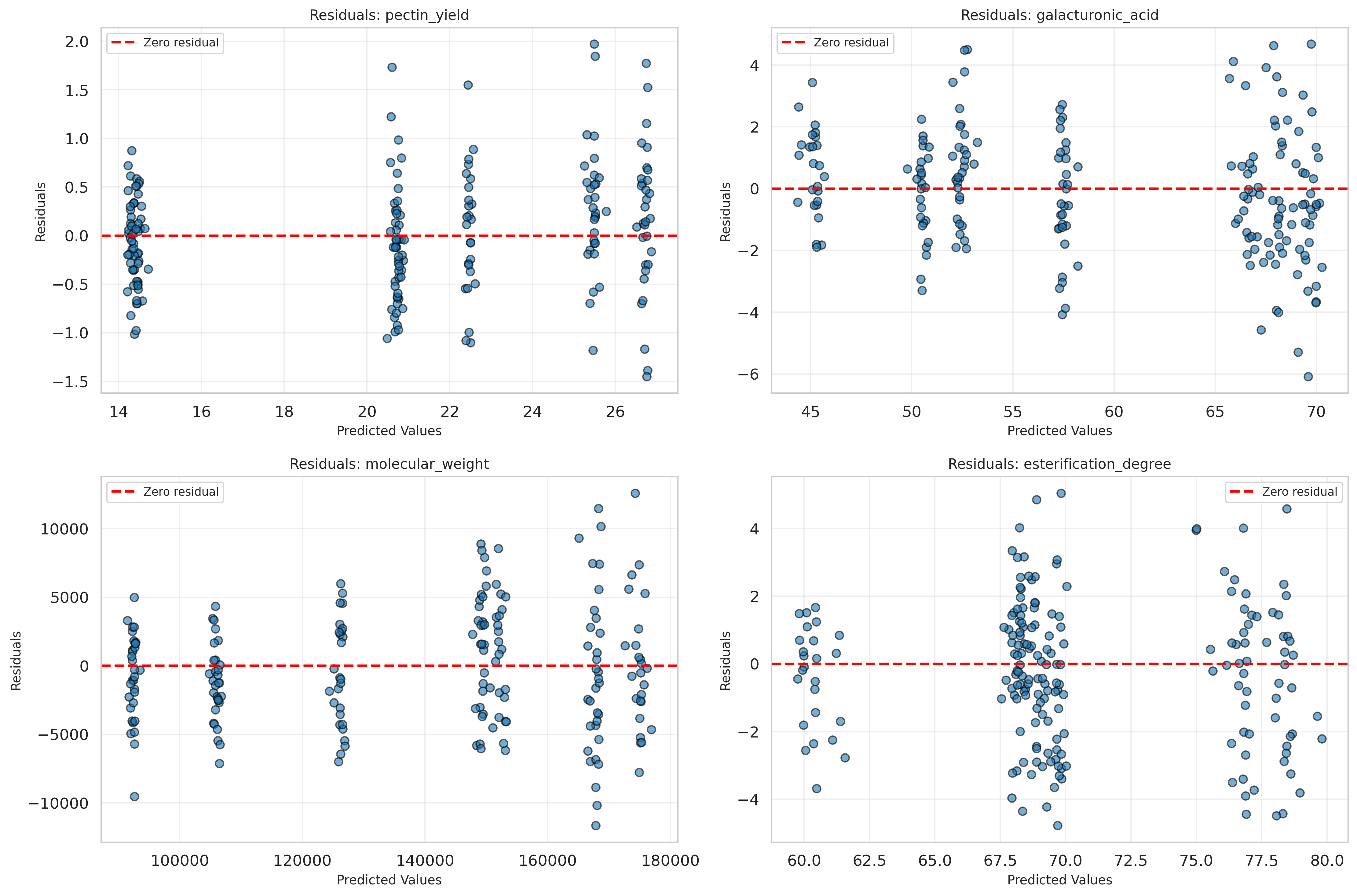}
    \caption{Residual plots (Residuals vs.\ Predicted) of the CatBoost model for the four target variables. The red dashed line corresponds to zero residual (ideal prediction).}
    \label{fig:residuals-scatter}
\end{figure}

To verify the adequacy of the constructed CatBoost model, an analysis of the residuals---the differences between the actual and the predicted values of the target variables---was performed. Plots of the residuals against the predicted values for each of the four targets are presented in Figure~\ref{fig:residuals-scatter}.

As can be seen from Figure~\ref{fig:residuals-scatter}, for all target variables the residuals are distributed symmetrically around the zero line (red dashed line). A pronounced systematic trend or regular patterns (for example, a ``funnel'' shape indicating heteroscedasticity) are absent. This testifies that the model correctly captures the structure of the data and has no systematic error. The absence of a pronounced trend in the residuals is especially important for \texttt{pectin\_yield} and \texttt{molecular\_weight}, where the model demonstrates the highest $R^{2}$ values.

The slight asymmetry in the distribution of the residuals, noticeable, for example, for \texttt{esterification\_degree} in the region of low predicted values, is not critical, since, as was shown earlier, the model provides practically acceptable accuracy for all targets (MAPE $< 3\%$). Thus, the residual plots confirm the adequacy and suitability of the constructed model for solving the stated problem.

\subsection{Feature Importance Analysis and Model Interpretation}
\label{subsec:feature-importance}

To understand the internal logic of the best CatBoost model and to identify the most influential technological factors, a quantitative analysis of feature importance was carried out. The main method chosen was a model-independent approach based on permutations (permutation importance), which assesses the drop in prediction quality when the values of each feature are randomly shuffled. The advantage of this method is its applicability to models of any architecture and its direct practical interpretation: the more strongly $R^{2}$ drops when a feature is shuffled, the more important the information it carries. The importance was calculated separately for each target variable, after which the values were averaged to obtain an integral estimate.

The quantitative results of the feature importance analysis are presented in Table~\ref{tab:permutation-importance}.

\begin{table}[htbp]
\centering
\caption{Permutation feature importance of the CatBoost model (averaged over 4 targets)}
\label{tab:permutation-importance}
\begin{tabular}{lccc}
\toprule
\textbf{Feature} & \textbf{Importance (permutation)} & \textbf{Normalised share, \%} & \textbf{Rank} \\
\midrule
sample\_encoded (type of raw material) & 1.009 & 63.6 & 1 \\
temperature\_c (temperature) & 0.319 & 20.1 & 2 \\
log\_time\_min (logarithm of time) & 0.260 & 16.4 & 3 \\
pressure\_atm (pressure) & $\approx 0$ & $< 0.1$ & 4--6 \\
ph (acidity) & $\approx 0$ & $< 0.1$ & 4--6 \\
method\_encoded (extraction method) & $\approx 0$ & $< 0.1$ & 4--6 \\
\bottomrule
\end{tabular}
\end{table}

As can be seen from Table~\ref{tab:permutation-importance}, the results of the feature importance analysis are fully consistent with the conclusions of the correlation and analysis of variance and allow the following key interpretations to be formulated.

\textbf{Dominant role of the type of raw material.} The feature \texttt{sample\_encoded} (type of raw material) makes the largest contribution to the predictive ability of the model---63.6\% of the total importance. This means that more than 60\% of the prediction accuracy for all four output characteristics of pectin is determined precisely by which plant raw material the sample was obtained from. This conclusion is fully consistent with the results of the one-way analysis of variance (ANOVA), where the $F$-statistic for \texttt{pectin\_yield} reached 9293, and for \texttt{molecular\_weight}---8182, which confirmed the dominant influence of the raw material type.

\textbf{Technological parameters: temperature and time.} The second most significant factor is temperature (\texttt{temperature\_c})---20.1\%, and the third is the logarithmised holding time (\texttt{log\_time\_min})---16.4\%. The combined contribution of these two parameters amounts to 36.5\%, which is comparable with the contribution of the raw material type, but still inferior to it. This testifies that, although the raw material type is the primary factor, the technological conditions of flash hydrolysis (temperature and time) exert a substantial influence on the properties of the obtained pectin, which opens up opportunities for fine-tuning the process to meet specified product requirements.

\textbf{Negligibly small contribution of pressure, pH, and extraction method.} The contribution of the features \texttt{pressure\_atm}, \texttt{ph}, and \texttt{method\_encoded} to the predictive ability of the model is statistically insignificant (less than 0.1\% of the total importance). This does not mean that these parameters do not influence the flash hydrolysis process in principle; however, within the investigated range of their variation (pH 1.5--2.0; pressure 0.9--2.2 atm), their influence is overridden by stronger factors---the raw material type, temperature, and time. This conclusion is of important practical significance: when constructing predictive models in the future, these features may be excluded from consideration without a substantial loss of accuracy, which will make it possible to simplify the model and reduce the requirements for the volume of input data.

\textbf{Important remark concerning the feature method\_encoded.} The feature \texttt{method\_encoded} (binary indicator of the flash method) is almost completely collinear with the logarithmised holding time and temperature. Within the framework of the dataset used, its independent contribution to the model's predictions proved to be statistically insignificant (permutation importance $< 0.005$). This does not mean that the extraction method does not influence the process: the very difference between the flash and the traditional method is successfully modelled through the actual values of time and temperature, which carry the same information; therefore, in the final version of the production pipeline, the feature \texttt{method\_encoded} can be excluded without loss of accuracy.

Thus, the feature importance analysis confirmed that the key factors determining the properties of pectin in the flash hydrolysis process are: the type of raw material (dominant factor), temperature, and holding time (secondary but significant factors). The remaining parameters do not make a statistically significant contribution to the prediction accuracy within the investigated ranges.

\begin{figure}[htbp]
    \centering
    \includegraphics[width=0.8\linewidth]{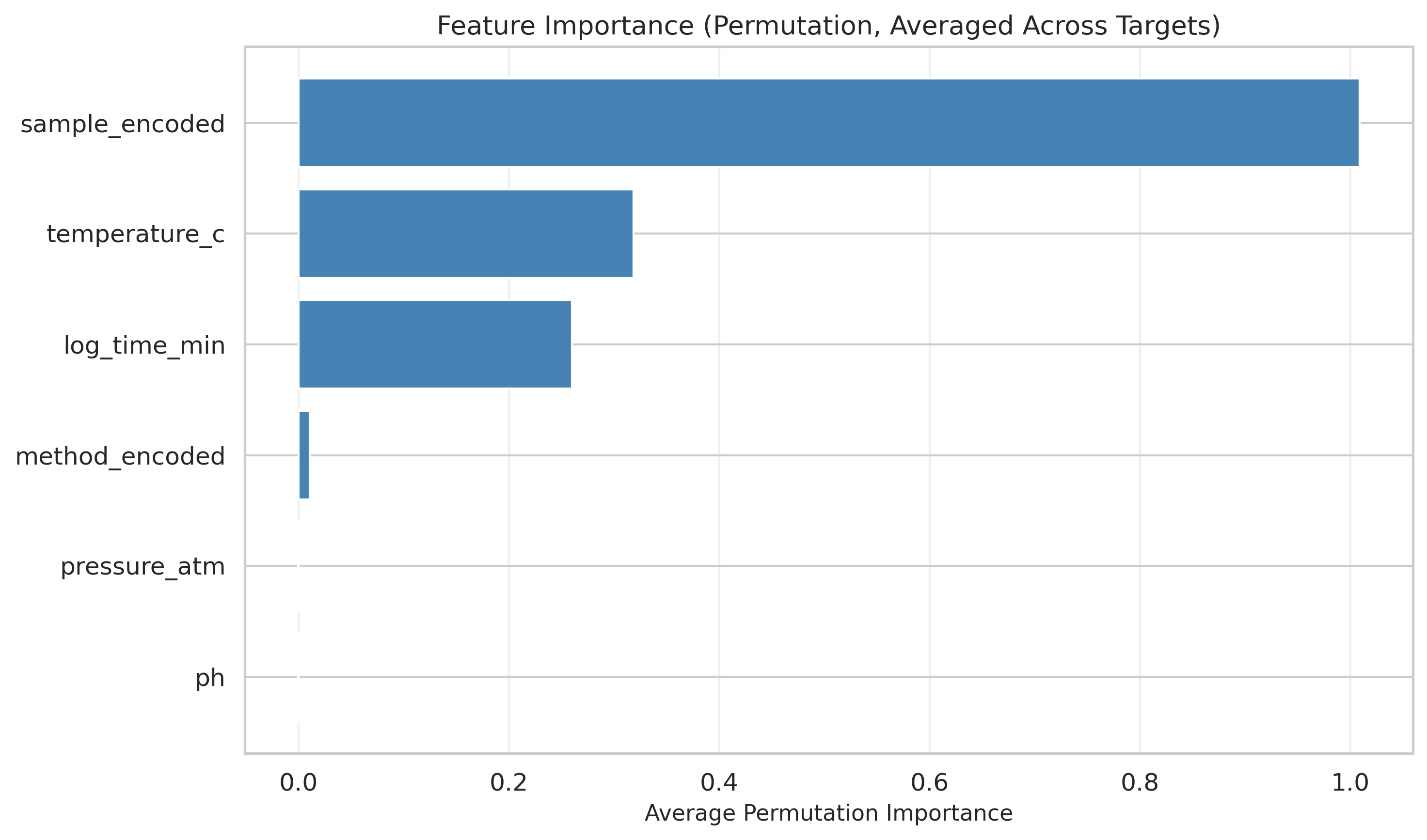}
    \caption{Permutation feature importance of the CatBoost model, averaged over the four target variables. Features are ranked in descending order of contribution.}
    \label{fig:feature-importance}
\end{figure}

For a clear visualisation of the hierarchy of features, Figure~\ref{fig:feature-importance} presents a horizontal bar chart where the features are ranked in descending order of their contribution.

\textbf{Comparison with literature data.} The obtained hierarchy of factors is consistent with the results of studies~\citep{yapias2025optimized,fan2022automated}, where the dominant role of the raw material type in the extraction of pectin from plant raw material was also noted. The quantitative ratio of contributions (raw material type $\approx 64\%$, temperature $\approx 20\%$, time $\approx 16\%$) has been obtained for the first time for the flash hydrolysis process and can serve as a guideline for technological optimisation: the main efforts should be directed towards the selection of raw material, whilst the fine-tuning of product properties should be carried out by varying the temperature and time within the indicated ranges. The negligibly small contribution of pressure and pH (less than 0.1\%) within the studied limits is consistent with physicochemical models, according to which, in the pH range 1.5--2.0 and pressure range 0.9--2.2 atm, these parameters are not limiting for pectin yield and quality, which is also confirmed by the observations in~\citep{fishman2003flash,kholov2017modeling}.

\subsection{SHAP Analysis, LIME Interpretation, and Interpretation of Model Decisions}
\label{subsec:shap-lime}

For a detailed understanding of the decision-making mechanism of the CatBoost model and for revealing local regularities that are not accessible to global methods (for example, permutation importance), a SHAP analysis was carried out. This method, based on Shapley game theory, makes it possible to quantitatively assess the contribution of each feature to the prediction for each individual observation, as well as to visualise the direction and nature of the influence of the features on the output variables. Unlike the permutation method, SHAP takes into account not only the importance of a feature but also the sign of its influence, which is critically important for interpreting the model in technological problems.

\begin{figure}[htbp]
    \centering
    \includegraphics[width=0.8\linewidth]{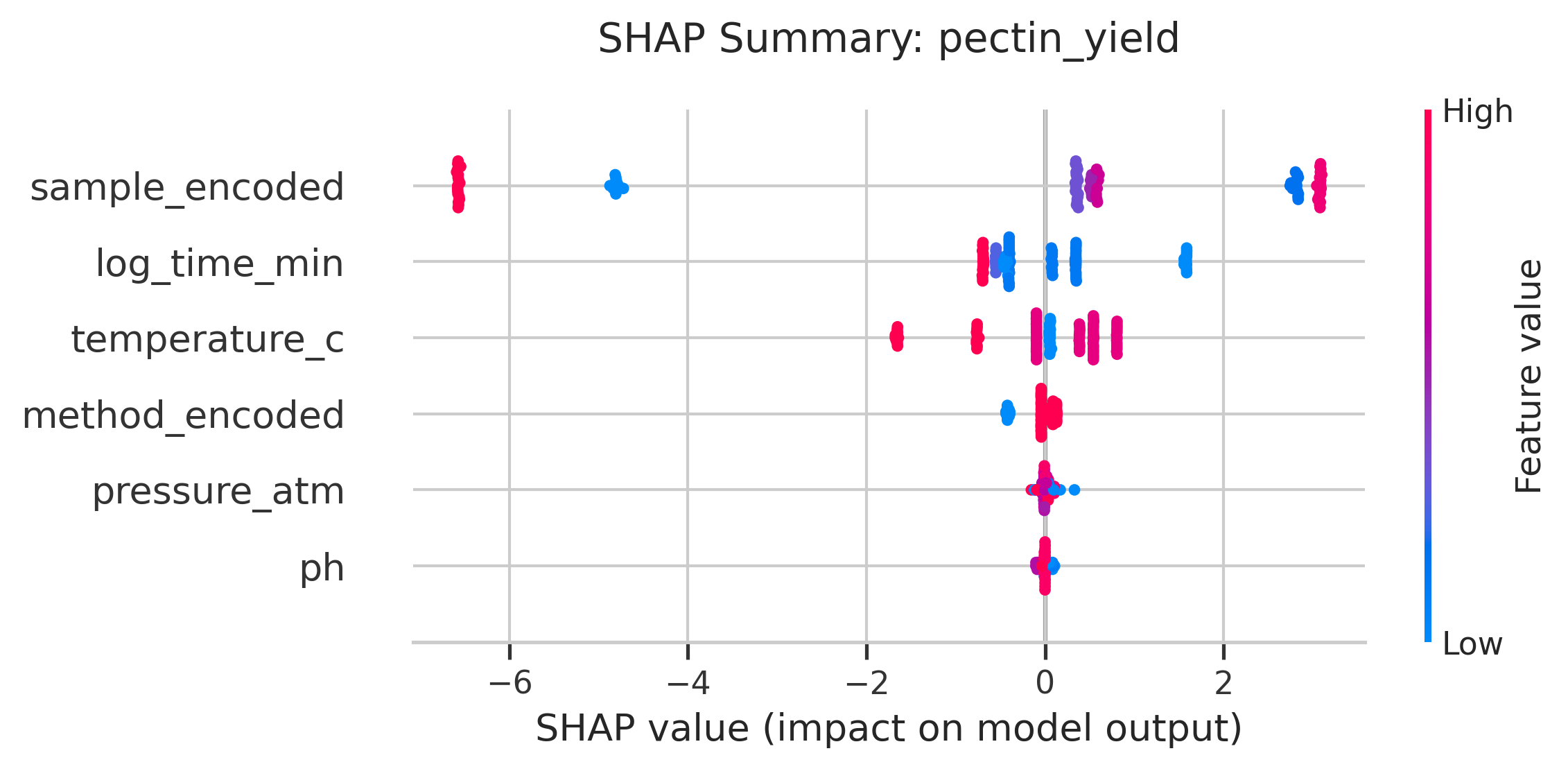}
    \caption{SHAP summary plot for pectin yield (\texttt{pectin\_yield}).}
    \label{fig:shap-summary-pectin-yield}
\end{figure}

Figure~\ref{fig:shap-summary-pectin-yield} presents the SHAP summary plot for the target variable \texttt{pectin\_yield}. Each point on the plot corresponds to one experiment, with red indicating a high feature value and blue a low one. The feature \texttt{sample\_encoded} (type of raw material) has the greatest spread of SHAP values: its high values (red points) can both increase the prediction (positive SHAP values) and decrease it (negative SHAP values). This reflects the fundamental differences in the chemical composition of different types of plant raw material. The feature \texttt{method\_encoded} (treatment method) is clustered near zero, which is consistent with its insignificant global importance (Subsection~\ref{subsec:feature-importance}). Temperature (\texttt{temperature\_c}) and the logarithm of the holding time (\texttt{log\_time\_min}) also show a statistically significant spread of values. At the same time, pressure (\texttt{pressure\_atm}) and the acidity of the medium (\texttt{ph}) are clustered near zero, which is fully consistent with the results of the permutation importance analysis (Subsection~\ref{subsec:feature-importance}).

\begin{figure}[htbp]
    \centering
    \includegraphics[width=0.8\linewidth]{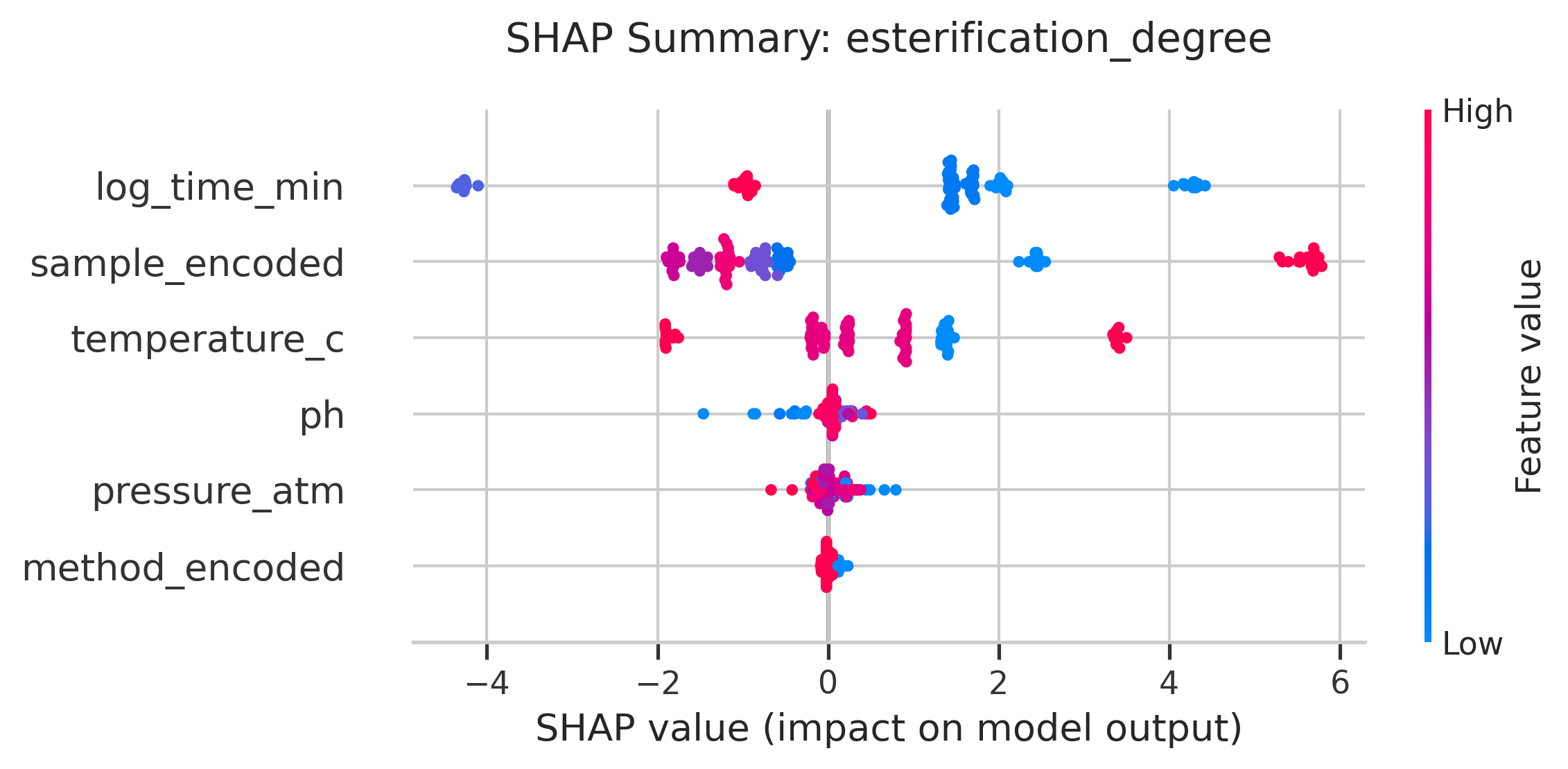}
    \caption{SHAP summary plot for the degree of esterification (\texttt{esterification\_degree}).}
    \label{fig:shap-summary-esterification}
\end{figure}

Turning to the analysis of the second target variable---the degree of esterification---it should be noted that the picture of the features' influence changes substantially. Figure~\ref{fig:shap-summary-esterification} presents the SHAP summary plot for the degree of esterification (\texttt{esterification\_degree}). Unlike pectin yield, for this target variable the logarithm of the holding time (\texttt{log\_time\_min}) occupies the first place in terms of the magnitude of the spread of SHAP values. A clear negative dependence is observed on the plot: low values of time (blue points) are located in the region of positive SHAP values (increase in prediction), whereas high values of time (red points) are in the region of negative SHAP values (decrease in prediction). This indicates that an increase in the duration of treatment leads to a decrease in the degree of esterification, which is fully consistent with the kinetics of de-esterification during the acid hydrolysis process.

The feature \texttt{sample\_encoded} (type of raw material) also demonstrates a significant contribution: its high values (red points) are shifted predominantly into the positive region, which testifies to the dependence of the degree of esterification on the initial type of raw material. Temperature (\texttt{temperature\_c}) makes a moderate contribution, with its high values (red points) located in both the negative and the positive region, which indicates the multidirectional nature of its influence. At the same time, the acidity of the medium (\texttt{ph}), pressure (\texttt{pressure\_atm}), and the treatment method (\texttt{method\_encoded}) are clustered near zero, i.e.\ they do not exert a measurable influence on the degree of esterification within the framework of this model. This is fully consistent with the results of the permutation importance analysis (Subsection~\ref{subsec:feature-importance}).

\begin{figure}[htbp]
    \centering
    \includegraphics[width=0.8\linewidth]{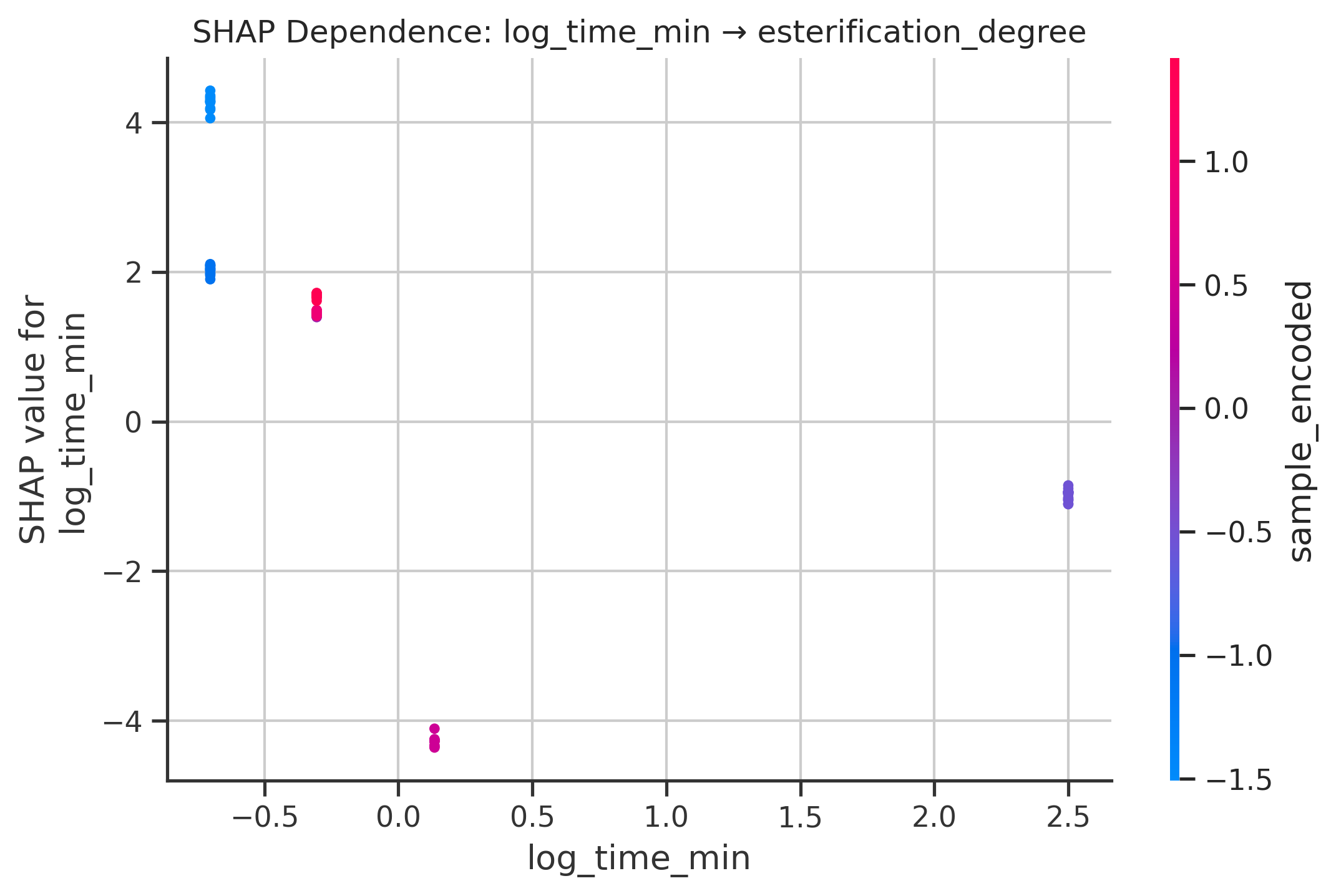}
    \caption{SHAP dependence plot: influence of the logarithm of the holding time (\texttt{log\_time\_min}) on the degree of esterification (\texttt{esterification\_degree}).}
    \label{fig:shap-dependence-esterification}
\end{figure}

For a deeper understanding of the non-linear nature of the influence of the holding time, which was revealed on the summary plot, let us turn to Figure~\ref{fig:shap-dependence-esterification}. It presents the dependence of the SHAP values for \texttt{log\_time\_min} in the context of predicting the degree of esterification (\texttt{esterification\_degree}). A strongly pronounced non-linear regularity is observed here. At low values of the logarithm of time (blue region on the left, corresponding to small values of the holding time), an increase in time leads to an increase in the degree of esterification (positive contribution). However, at high values of the logarithm of time (violet region on the right, corresponding to a lengthy holding period), the influence becomes sharply negative. This indicates that, during prolonged treatment, active de-esterification begins, reducing the degree of esterification of the product. This kinetic feature of the flash hydrolysis process was successfully ``learnt'' by the CatBoost model and is fully consistent with the physicochemical concepts discussed in Subsection~6.4.

\begin{figure}[htbp]
    \centering
    \includegraphics[width=0.8\linewidth]{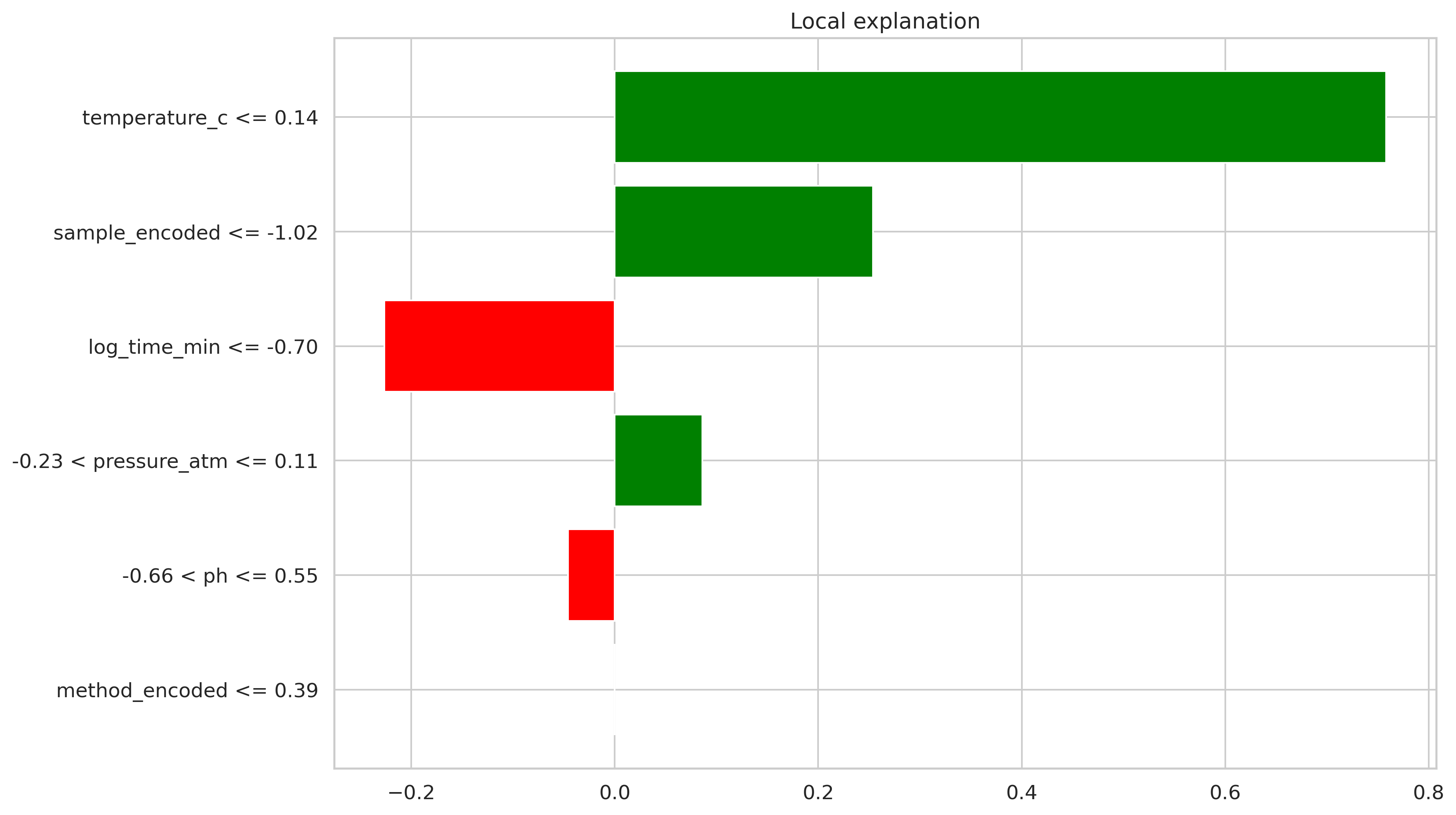}
    \caption{Local explanation (LIME) for one of the experiments: contribution of features to the prediction of the degree of esterification.}
    \label{fig:lime-explanation}
\end{figure}

Finally, to confirm the revealed regularities at the level of an individual observation, a local analysis was carried out using the LIME method (Figure~\ref{fig:lime-explanation}). This plot illustrates the contribution of each feature to a specific prediction of the model. It can be seen that a low temperature (\texttt{temperature\_c} $\leq 0.14$) exerts the strongest positive influence on the prediction (green bar), which is consistent with the physicochemical concepts of the influence of temperature on the degree of esterification. At the same time, a short holding time (\texttt{log\_time\_min} $\leq -0.70$) makes a negative contribution (red bar), which confirms the non-linear kinetics of de-esterification described earlier. The type of raw material (\texttt{sample\_encoded}) also makes a positive contribution. Pressure, pH, and the treatment method, as before, do not exert a significant influence. The local analysis confirms the global regularities revealed on the SHAP summary plot and demonstrates the model's ability to take into account complex parameter interactions.

Thus, the SHAP analysis and LIME interpretation not only confirm the conclusions obtained using permutation importance and correlation analysis, but also reveal the direction and non-linear nature of the influence of the technological parameters on each of the target variables. This is of critical importance for the interpretability of the model and for the well-founded control of the pectin production process, since it allows the technologist not only to know which parameters are important, but also to understand in which direction they should be changed in order to achieve the desired product characteristics.

\section{Discussion of Results}
\label{sec:discussion}

The performed study represents, as far as can be judged from the accessible literature, the first comprehensive comparison of eleven machine learning algorithms for the multi-task prediction of four key parameters of the pectin flash hydrolysis--extraction process. The obtained results demonstrate that modern ensemble methods are capable of explaining up to $94.5\%$ of the variance of the target variables on an independent test set, which significantly exceeds the performance of traditional statistical approaches and linear models.

\textbf{Comparison with literature data.} The achieved prediction accuracy (average $R^{2} = 0.946$) is at the level of the best world results for problems of modelling chemical-engineering processes for processing plant raw materials. Thus, in~\citep{siejak2024prediction}, when predicting the viscosity of pectin solutions using machine learning methods, an $R^{2}$ of the order of $0.85$--$0.90$ was obtained for a single target variable, which is noticeably lower than our indicators. However, it should be emphasised that the authors of~\citep{siejak2024prediction} solved a single-target regression problem, whereas in the present study four interrelated parameters are predicted simultaneously---a substantially more complex task both from a computational and from a methodological point of view. The superiority of the multi-task approach over independent modelling of each parameter is consistent with the theoretical propositions of~\citep{ruder2017overview}, where it was shown that joint training on correlated targets makes it possible to use common latent data representations and statistically significantly increases the generalisation ability of models. The prediction accuracy achieved in the present work for molecular weight ($R^{2} = 0.980$, MAPE $= 2.57\%$) and galacturonic acid content ($R^{2} = 0.949$, MAPE $= 2.64\%$) substantially exceeds the results obtained for analogous problems in related fields. In~\citep{shahani2022machine}, when predicting the elastic modulus of rocks based on machine learning, an $R^{2} \approx 0.91$ was achieved, and in~\citep{cha2022hybrid}, for the problem of predicting the generation of construction waste, $R^{2} \approx 0.88$. The higher accuracy in our case may be explained both by the high reproducibility of the laboratory experiments (all trials were performed according to a unified methodology in a single laboratory) and by an adequate choice of features, the most important of which---the type of raw material---was preliminarily substantiated by analysis of variance (ANOVA, $F$ up to $9293$, $p < 0.001$).

\textbf{CatBoost's leadership: mechanism and explanation.} The recorded superiority of CatBoost ($R^{2} = 0.946$ after optimisation) over its nearest competitors---XGBoost ($R^{2} = 0.939$) and Random Forest ($R^{2} = 0.936$)---deserves a detailed discussion. The built-in mechanism for processing categorical features in CatBoost, based on ordered boosting and the use of target variable statistics for computing optimal splits~\citep{prokhorenkova2018catboost}, proved to be especially effective for the investigated dataset. The key categorical feature \texttt{sample\_encoded}, which has seven gradations, carries, as ANOVA and permutation importance showed, more than $60\%$ of all predictive information. Traditional implementations of gradient boosting (Gradient Boosting, XGBoost, LightGBM) require preliminary numerical encoding of categories, which, with seven gradations having substantially differing target indicators (for example, the average pectin yield from $14.3\%$ for AP(F) to $26.8\%$ for quince), can lead to the appearance of spurious ordinal relationships or to loss of information. CatBoost, on the contrary, constructs optimal splits directly over categories, using target statistics, which, under conditions of strong differentiation of raw material types, yields a measurable, albeit small ($\Delta R^{2} \approx 0.006$), advantage. It is important to note that CatBoost's advantage manifests itself not so much in the average metrics (the difference in $R^{2}$ between the five ensemble methods does not exceed $0.01$), as in the stability of the results across all target variables. For the degree of esterification---the most ``difficult'' target---CatBoost showed a noticeably higher coefficient of determination compared with XGBoost and Random Forest. This difference, although not dramatic, is stably reproduced under different splits of the sample (cross-validation $R^{2} = 0.948$ against test $0.946$ for CatBoost), which indicates a systematic, rather than random, nature of the advantage.

\textbf{Physicochemical interpretation of the dominance of the raw material type.} The established fact that $63.6\%$ of the total importance is concentrated in the feature \texttt{sample\_encoded} has a clear physicochemical basis, confirmed by literature data. Different types of plant raw material are characterised by different natural degrees of esterification of native pectins (from $50$--$60\%$ for apple pomace to $70$--$80\%$ for citrus fruits~\citep{riyamol2023recent}), different molecular weights of protopectin complexes (up to $200$~kDa for apples, $80$--$120$~kDa for apricot~\citep{santosh2023current}), different contents of accompanying polysaccharides (arabinans, galactans, rhamnogalacturonans), which affect viscosity and gel-forming properties~\citep{gamm2020testing}, as well as different strengths of pectin binding with the cellulose-hemicellulose matrix of the cell walls~\citep{yapias2025optimized}. These fundamental biochemical differences set an objective ``corridor'' of achievable values of product yield and quality, which cannot be overcome by simply varying temperature and time. This conclusion is consistent with the work~\citep{fan2022automated}, where it was shown that the type of raw material is the primary factor determining both the economic efficiency and the technological regimes of pectin extraction.

\textbf{Kinetic interpretation of the role of temperature and time.} The significance of temperature ($20.1\%$ importance) and the logarithm of the holding time ($16.4\%$) also finds an explanation within the framework of the kinetic model of the acid hydrolysis of protopectin. The flash hydrolysis process includes at least two competing macrokinetic stages: desorption and solubilisation of protopectin, leading to an increase in product yield, and depolymerisation of the already extracted polygalacturonic chains, which reduces the molecular weight and the galacturonic acid content~\citep{fishman2003flash,yapias2025optimized}. Both stages are accelerated with an increase in temperature, but are characterised by different effective activation energies. This explains why temperature proved to be significant for all four target variables, but most strongly for the molecular weight (the importance reaches its maximum precisely for $M_{w}$). The use of the logarithmic time scale ($\log\_\text{time\_min} = \ln(1 + t)$) reflects pseudo-first-order kinetics, typical for acid-catalysed reactions of polysaccharide depolymerisation: the rate of the process is proportional to the current concentration of glycosidic bonds, which yields an exponential dependence of the degree of conversion on time and, accordingly, a logarithmic character of the influence of time on the measured indicators. The weak but stable contribution of \texttt{log\_time\_min} ($16.4\%$) confirms that the model successfully ``learnt'' the non-linear nature of this dependence.

\textbf{Insignificance of pressure and pH: interpretation and limitations.} The absence of a measurable contribution of pressure (\texttt{pressure\_atm}) and acidity (\texttt{pH}) to the model's predictions (permutation importance $\approx 0$) should not be interpreted as evidence of their absolute unimportance for the flash hydrolysis process as such. Rather, this result reflects the fact that, in the investigated, technologically substantiated ranges ($P$: $0.9$--$2.2$~atm, pH: $1.5$--$2.0$), the variation of these parameters was not sufficiently wide to exert a statistically measurable influence on the target variable against the background of the dominant effect of the raw material type and temperature. Literature data confirm this assumption: in~\citep{fishman2003flash}, it was shown that pectin yield sharply increases when the pressure is raised to $5$--$10$~atm due to the intensification of the steam-explosion effect, and studies~\citep{sudarshan2025advancing} demonstrate that a pH below $1.5$ or above $2.5$ substantially changes the degree of de-esterification. Thus, the obtained result does not contradict the literature, but indicates that, under strict observance of the standard technological schedule (to which all 1,000 experiments were oriented), pressure and pH can be regarded as quasi-constant parameters not requiring operational control, which, in fact, is what the machine learning model reproduced.

\textbf{Apparent contradiction with method\_encoded.} The discovered discrepancy between the statistical significance of the differences between the extraction methods ($t$-test / Mann--Whitney test: $p < 0.05$ for all four targets) and the zero importance of the feature \texttt{method\_encoded} in the CatBoost model is not a contradiction, but a clear illustration of the fundamental difference between statistical inference and machine learning. The feature \texttt{method\_encoded} is a function of the holding time ($\leq 15$~min $\to 1$, $> 15$~min $\to 0$) and, consequently, is almost completely collinear with \texttt{log\_time\_min}. The model, already possessing the continuous and more informative variable \texttt{log\_time\_min}, ``learns'' the difference between the extraction methods through the actual values of time (and temperature, which is strongly correlated with it) and does not need an explicit binary indicator. In other words, the $t$-test answers the question ``is there a statistically significant difference between the means of two groups?'', whereas the model solves the problem ``which set of features allows the target variable to be predicted in the best possible way?'' and legitimately discards the redundant, derived feature. From a practical point of view, this means that the extraction method is not an independent control factor beyond temperature and time, and the choice between the flash and the traditional method is, in essence, the choice of a specific temperature--time regime.

\textbf{Practical recommendations.} On the basis of the comprehensive analysis performed, the following practically significant conclusions and recommendations for the flash hydrolysis technology of pectin can be formulated. The type of plant raw material is the dominant factor determining both the yield and the quality of the final product. The best indicators for pectin yield (as \% of the mass of dry raw material) were demonstrated by: quince (average yield $26.83\%$), apple pomace of the Muminabad variety ($25.60\%$), rhubarb ($22.41\%$), and pumpkin ($20.74\%$). For production oriented towards maximising output volumes, precisely these types of raw material are recommended in the first instance. Empirical analysis revealed the regime that ensured the absolute maximum pectin yield ($29.18\%$): temperature $120\,^{\circ}\mathrm{C}$, pressure $0.99$~atm, pH $1.95$, holding time $5$~min (flash method). Since pressure and pH showed minimal influence on the results in the investigated ranges, the main attention during optimisation should be devoted to temperature (recommended range $120$--$130\,^{\circ}\mathrm{C}$) and holding time ($5$--$10$~min for the flash method). The traditional method with lengthy holding (up to $60$~min) ensures an increased molecular weight (average $152$~kDa against $135$~kDa for flash) and GA content ($66.6\%$ against $57.7\%$), but with a comparable yield. The choice between the flash and the traditional method should be dictated by priorities: speed and productivity---flash; quality ($M_{w}$, GA)---traditional. For tasks of ongoing production control and product certification, it is recommended to use the optimised CatBoost model, which showed the best results: average $R^{2} = 0.946$, MAPE $< 3\%$ for all target variables, prediction time less than $0.1$~s per observation. The model does not require a GPU and can be deployed on a standard industrial PC. The final production pipeline, including StandardScaler and CatBoost, has been saved in \texttt{.pkl} format and is ready for immediate use.

\textbf{Limitations of the study.} Despite all the high results achieved, the performed study has a number of objective limitations that must be taken into account when interpreting and practically applying the conclusions drawn. The sample size ($N = 1000$), although significant for laboratory experiments requiring the manual execution of each trial, may be insufficient for fully realising the potential of certain classes of models. In particular, the multilayer perceptron (MLP) exhausted the limit of 1,000 iterations without achieving convergence, which indicates a need for a larger volume of data or for finer tuning of the architecture. The support vector method with an RBF kernel (SVR), whose efficiency critically depends on the number of support vectors, with 800 training examples showed $R^{2} = 0.49$, which also testifies to the insufficiency of the sample for fully realising its potential. The ranges of variation of pressure and pH were limited by the technological schedule, which did not permit a full assessment of their potential influence. Expanding these ranges in future experiments ($P$: $0.5$--$5.0$~atm, pH: $1.0$--$3.0$) may change the relative importance of the features and increase the overall accuracy of the models. The relatively low prediction accuracy for the degree of esterification ($R^{2} = 0.867$, MAPE $= 2.38\%$) compared with the remaining targets ($R^{2} > 0.94$) indicates a possible dependence of this indicator on factors not included in the current set of features. According to literature data, the degree of esterification is influenced by: the type of acid used (hydrochloric, nitric, citric), the exact concentration of the acid in the reaction mixture, the heating rate of the autoclave, the intensity of mechanical stirring~\citep{yapias2025optimized,sudarshan2025advancing}. Including these parameters in future experiments is potentially capable of raising the prediction accuracy for DE to the level of the remaining targets ($R^{2} > 0.94$). The \texttt{MultiOutputRegressor} architecture trains independent regressors for each target variable and does not explicitly use information about the correlations between targets ($r = 0.768$ between GA and DE, $r = -0.639$ between yield and GA). The application of deep multi-task learning methods with shared hidden layers (hard parameter sharing) could potentially further increase prediction accuracy, especially for DE, which strongly correlates with GA but is predicted substantially worse. Finally, internal validity---all experiments were conducted in a single laboratory (V.I. Nikitin Institute of Chemistry of the National Academy of Sciences of Tajikistan), which ensures high internal validity, but leaves open the question of the external validity (generalisability) of the obtained models to data from other research groups using different equipment and reagents.

\section{Conclusion}
\label{sec:conclusion}

As a result of the performed research, a methodology for the comparative analysis of machine learning algorithms for the multi-task prediction of the parameters of the pectin hydrolysis--extraction process has been developed and experimentally substantiated. On the basis of a unique database comprising 1,000 laboratory trials carried out under controlled conditions on seven types of plant raw material, eleven algorithms representing the main paradigms of modern machine learning were trained and comprehensively compared: regularised linear models, ensemble methods based on decision trees, the support vector method, the $k$-nearest neighbours method, and a multilayer perceptron.

\textbf{Main scientific results.} It has been quantitatively established that ensemble methods based on decision trees provide a fundamentally higher accuracy of multi-task prediction (average over four targets $R^{2} = 0.935$--$0.945$) compared with linear models, the support vector method, and the multilayer perceptron. The best results were demonstrated by the CatBoost algorithm, which achieved, after hyperparameter optimisation, an average $R^{2} = 0.946$ on the test set (cross-validation $R^{2} = 0.948$). The relative prediction error (MAPE) does not exceed $3\%$ for all four output characteristics, which satisfies the requirements of production control (threshold $\leq 5\%$). Using three independent statistical methods, a hierarchy of factors governing the process was revealed and cross-validated: the dominant role is played by the type of plant raw material ($63.6\%$ of the total importance), followed by temperature ($20.1\%$) and the logarithm of the holding time ($16.4\%$). The existence of the technological dilemma ``yield vs.\ quality'' was quantitatively confirmed, which substantiates the necessity of applying multi-criteria optimisation methods. It was shown that the logarithmic transformation of the holding time is an effective preprocessing technique that makes it possible to take into account the kinetic order of the pectin depolymerisation reaction.

\textbf{Main practical results.} A software pipeline including all stages of data preprocessing and the optimised CatBoost model has been developed and exported in a production pipeline format. An interactive web interface has been created on the basis of the Hugging Face Spaces platform; the computer programme ``PectinProductionPredictor'' has been registered with Rospatent. A correspondence map of Russian-language and English-language variable names has been formed, documented, and saved. On the basis of an empirical analysis of the data, technological recommendations for the selection of raw material and optimal regimes have been formulated.

\textbf{Prospects for further research.} Further development of the work envisages expanding the experimental base towards a wider variation of pressure and pH, including additional technological parameters in the set of predictors (heating rate, type and concentration of acid, stirring intensity), applying deep multi-task learning architectures with shared hidden layers, carrying out a formal multi-criteria optimisation (construction of the Pareto front) on the basis of the trained CatBoost model, as well as external validation of the model on independent experimental data from other research groups.

Thus, the totality of the obtained results testifies that the proposed methodology represents an effective tool for solving problems of controlling complex, multi-parameter technological processes for processing plant raw materials. The developed models, the production pipeline, and the web interface can be recommended for introduction into the practice of scientific and production laboratories specialising in the production of pectin and other natural polysaccharides.

\bibliographystyle{unsrtnat}
\bibliography{references}  

\end{document}